\documentclass[3p,times]{elsarticle}

\usepackage{ecrc}

\volume{00}

\firstpage{1}

\journalname{Information Sciences}

\runauth{M.K. Lim et al.}

\jid{ins}

\jnltitlelogo{Information Sciences}

\usepackage{amssymb}

\usepackage{graphicx,color}
\usepackage{algorithmic, subfigure}
\usepackage{mdwmath,amsthm}
\usepackage[cmex10]{amsmath}
\usepackage{multirow}
\usepackage{enumerate}

\usepackage[figuresright]{rotating}

% ...

\begin{document}

\begin{frontmatter}


 \tnotetext[label1]{Corresponding Author: Mei Kuan Lim (email: imeikuan@siswa.um.edu.my; phone/fax: +0060379676433)}

\dochead{}

\title{Refined Particle Swarm Intelligence Method for Abrupt Motion Tracking}


\author{ Mei Kuan Lim$^1$,
Chee Seng Chan$^1$, Dorothy Monekosso$^2$
and Paolo Remagnino$^3$}

\address{$^1$University of Malaya, Center of Image and Signal Processing, 50603 Kuala Lumpur, Malaysia;\\ 
$^2$University of West England, Eng. \& Maths., Bristol BS16 1QY, United Kingdom; \\ 
$^3$Kingston University, Comp. \& Info. Sys., Surrey KT1 2EE, United Kingdom \\}

\begin{abstract}
Conventional tracking solutions are not feasible in handling abrupt motion as they are based on smooth motion assumption or an accurate motion model. Abrupt motion is not subject to motion continuity and smoothness. To assuage this, we deem tracking as an optimisation problem and propose a novel abrupt motion tracker that based on swarm intelligence - the SwaTrack. Unlike existing swarm-based filtering methods, we first of all introduce an optimised swarm-based sampling strategy to tradeoff between the exploration and exploitation of the search space in search for the optimal proposal distribution. Secondly, we propose Dynamic Acceleration Parameters (DAP) allow on the fly tuning of the best mean and variance of the distribution for sampling. Such innovating idea of combining these strategies in an ingenious way in the PSO framework to handle the abrupt motion, which so far no existing works are found. Experimental results in both quantitative and qualitative had shown the effectiveness of the proposed method in tracking abrupt motions.
\end{abstract}


\end{frontmatter}


\section{Introduction}
\label{intro}

Visual tracking is one of the most important and challenging research topics in computer vision. One of the main reason is due to it's pertinent in the tasks of motion based recognition, automated surveillance, video indexing, human-computer interaction and vehicle navigation \cite{Yang,Yilmaz}. In general, motion estimation in a typical visual tracking system can be formulated as a dynamic state estimation problem: $x_t=f(x_t-1,v_t-1)$ and $z_t = h(x_t,w_t)$, where $x_t$ is the current state, $f$ is the state evolution function, $v_t-1$ is the evolution process noise, $z_t$ is the current observation, $h$ denotes the measurement function, and $w_t$ is the measurement noise. The task of motion estimation is usually completed by utilizing predictors such as Kalman filters \cite{welch1995introduction,wan2000unscented,Oussalah200085} particle filters \cite{isard1998condensation,arulampalam2002tutorial,Liu2012141} or linear regression techniques \cite{ellis2011linear}. This is commonly further enhanced by assuming that motion is always governed by Gaussian distribution based on Brownian motion or constant velocity motion models \cite{Yilmaz, Garcia}.

\begin{figure}[ht]
\centering
\includegraphics[height=0.4\linewidth, width=0.85\linewidth]{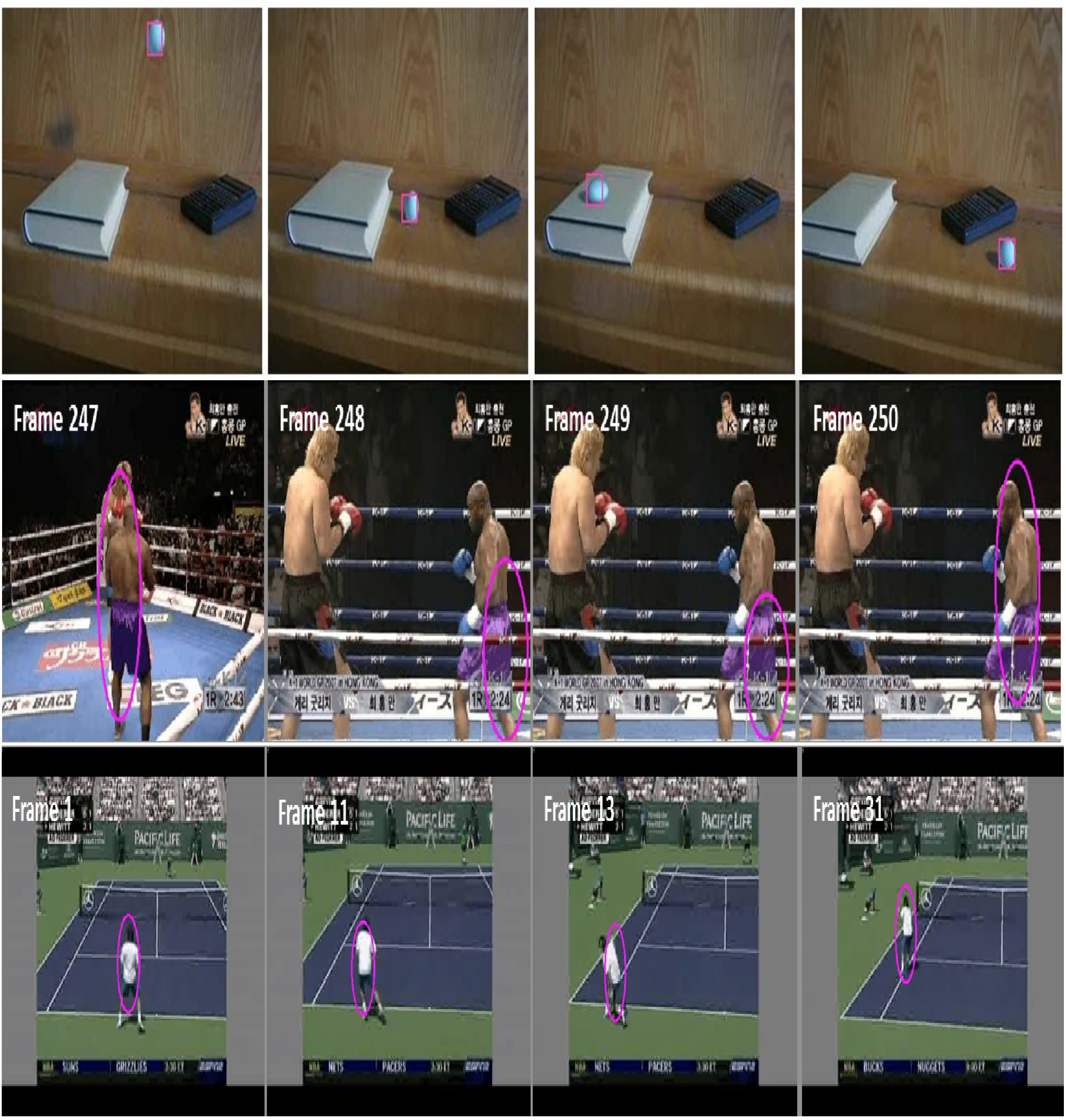}
\caption{Example of the abrupt motion in different scenarios. Top: Abrupt Motion with Inconsistent Speed. Middle: Switching of Camera during a Boxing Game. Bottom: Low Frame Rate of Video due to Downsampling.}
\label{fig:intro1}
\end{figure}

While this assumption holds true to a certain degree of smooth motion, it tends to fail in the case of abrupt motion such as fast motion (e.g. the movement of ball in sport events), camera switching (tracking of subject in a camera topology), low frame-rate video as illustrated in Fig. \ref{fig:intro1}. The main reason is that the state equation could not cope with the unexpected dynamic movement, e.g. sudden or sharp changes of the camera/object motion in adjacent frames. Nonetheless, such sampling-based solutions also suffered from the well-known local trap problem and particle degeneracy problem. In order to handle these issues, one of the earliest work \cite{Li2009} considered tracking in low frame rate video. The work considered tracking in low frame rate as to abrupt motion and proposed a cascade particle filters to tackle this issue. This is, then, followed by a number of sampling strategies \cite{Kwon2008,Kwon2010,Kwon2012,Zhang2010,Zhou} incorporated into Markov Chain Monte Carlo (MCMC) tracking framework. Their method alleviates the constant velocity motion constraint in MCMC by improvising the sampling efficiency. 

These aforementioned works have shown satisfactory results in tracking abrupt motion, however, we observed that most of the work had been focused on employing different sampling strategy into the Bayesian filtering framework. There are clear trends of increased complexity; as methods have gotten more complicated to cope with more difficult tracking scenarios. Often these sophisticated methods compensate the increased in complexity in a certain aspect of the algorithm by reducing the other aspect of it. For example, the increased number of subregions for sampling to cope with the variation of abrupt motion is compensated by using a smaller number of samples to reduce, if not maintaining the computational cost incurred. However, are these complex and sophisticated methods really necessary?

Recently, Particle Swarm Optimization (PSO) \cite{Eberhart,vandenBergh2006937,zhang2008sequential,Tong,Neri201396}, a new population based stochastic optimization technique, has received more and more attention because of its considerable success. Unlike the independent particles in the particle filter, the particles in PSO interact locally with one another and with their environment in analogy with the cooperative and social aspects of animal populations, for example as found in birds flocking. With this, Li et al. \cite{li2009contour} employed PSO in contour tracking problem to handle the abrupt motions. However, in the PSO method, there is a possibility that most samples will get trapped in a few strong local maxima. Hence, the PSO method fails to track highly abrupt motions.



In this paper, we proposed SwaTrack - Swarm intelligence-based Tracking algorithm to handle the abrupt motion. Our contributions are firstly, in contrast to the conventional abrupt motion solutions that based on different sampling methods in Bayesian filtering which are computational expensive, we deem tracking as an optimisation problem and adopted particle swarm optimisation algorithm soley as the motion estimator. In particular, we replace the state equation, $x_t=f(x_t-1,v_t-1)$ with a novel velocity model in the PSO. Secondly, we introduced Dynamic Acceleration Parameters (DAP) and Exploration Factor ($\mathcal{EF}$) into the PSO framework to avoid the swarm explosion and divergence problem in tracking highly abrupt motion. While the PSO is not new, it is the innovating idea of combining the DAP and $\mathcal{EF}$ in an ingenious way to handle the abrupt motion which so far no existing works are found. Experimental results using a large scale of public datasets and comparison with the state-of-the-art algorithms have shown the effectiveness and robustness of the proposed method in terms of dataset unbiased, different size of object and recovery from error.

The rest of this paper is organised as follows. In Section \ref{related}, we provide the background work in tracking abrupt motion. The PSO is revisited in Section \ref{sec:ParticleSwarm} and its limitation to handle abrupt motion. Proposed work is detailed in Section \ref{sec:ParticleSwarm} while experimental results and discussion are given in Section~\ref{sec:experiment}. Finally, conclusion is drawn in Section~\ref{sec:conclusion}.

\section{Related Work}
\label{related}

\begin{figure}[t]
\centering
\subfigure[Particle degeneracy]{\includegraphics[height=0.2\linewidth, width=0.4\linewidth]{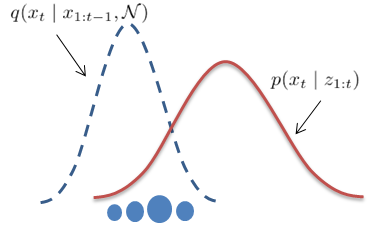}
\label{fig:pfprob01}}
\subfigure[Trapped in local optima]{\includegraphics[height=0.2\linewidth, width=0.4\linewidth]{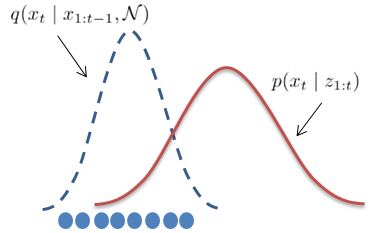}
\label{fig:pfprob02}}
\caption{Known problem of sampling-based tracking such as particle filter tracking and its variation (a) particle degeneracy problem (b) trapped in local optima.}
\label{fig:pfproblem}
\end{figure}

While considerable research efforts exist in relation to visual tracking, only a handful corresponds to abrupt motion \cite{Wong, Kwon2012, Liu, Zhou}. Abrupt motion can be defined as situations where the object’s motion changes at adjacent frames with unknown pattern in scenarios such as i) partially low-frame rate, ii) switching of cameras view in a topology network or iii) the irregular motion of the object. Therefore conventional sampling-based solutions that assume Gaussian distribution based on Brownian motion or constant velocity motion models tend to fail in this area as illustrated in Fig. \ref{fig:pfproblem} . For a complete review on the general visual tracking, we encourage the reader to \cite{Yang,Yilmaz} while this literature review section will only focus on work that handle abrupt motion. 

In the recent work, Markov Chain Monte Carlo (MCMC) has been proposed to overcome the computational complexity in PF when the state space increases \cite{Zuriarrain}. While MCMC methods cope better in a high-dimensional state space, a common problem is the need to have a large number of samples, especially when tracking abrupt motion. Thus, to deal with abrupt motion, there has been a handful of work which introduced modifications and refinements on the conventional MCMC. Kwon \emph{et al.} in \cite{Kwon2008}, introduced an integration of the Wang-Landau algorithm into the MCMC tracking framework to track abrupt motion. Their method alleviates the constant-velocity motion constraint in MCMC by improvising the sampling efficiency using the proposed annealed Wang-Landau Monte Carlo (A-WLMC) sampling method. The A-WLMC method increases the flexibility of the proposal density in MCMC by utilising the likelihood and density of states terms for resampling. Then, another variation of MCMC known as the interactive MCMC (IMCMC) was proposed \cite{Kwon2010}, where multiple basic trackers are deployed to track the motion changes of a corresponding object. The basic trackers which comprise of different combinations of observation and motion models are then fused into a compound tracker using the IMCMC framework. The exchange of information between these trackers has been shown to cope with abrupt motion while retaining the number of samples used. In another advancement, an intensively adaptive MCMC (IA-MCMC) sampler \cite{Zhou} has been proposed. Their method further reduces the number of samples required when tracking abrupt motion by performing a two-step sampling scheme; the preliminary sampling step to discover the rough landscape of the proposal distribution (common when there is large motion uncertainty in abrupt motion) and the adaptive sampling step to refine the sampling space towards the promising regions found by the preliminary sampling step. In another attempt for effective sampling of abrupt motion, \cite{Kwon2012} proposed the N-fold Wang-Landau (NFWL) tracking method that uses the N-fold algorithm to estimate the density of states which will then be used to automatically increase or decrease the variance of the proposal distribution. The NFWL tracking method copes with abrupt changes in both position and scale by dividing the state space into larger number of subregions. Therefore, the N-fold algorithm was introduced during sampling to cope with the exponentially increased number of subregions. 

Motivated by the meta-level question prompted in \cite{Zhu} on \emph{whether there is a need to have more training data or better models for object detection}, we raise similar question in the domain of this area; will continued progress in visual tracking be driven by the increased complexity of tracking algorithms? As indicated in the earlier section, often these sophisticated methods compensate the increased in complexity in a certain aspect of the algorithm by reducing the other aspect of it. Furthermore, according to \cite{Garcia}, different scenarios require different dynamic models. If \emph{motion models only work sometimes}, on a particular scenario, then how far should the increased in complexity of tracking algorithms be, in order to cope with the challenges of real-time tracking scenarios? Should we look into less complex methods instead, since motion models only work sometimes? Hence, we study a simple and yet effective algorithm, the SwaTrack that utilise the PSO framework to effectively handle the abrupt motion using the particles sharing information themselves. We deem the tracking as an optimisation problem, and hence the proposed method is dataset unbias and able to recover from error. 

Work that are considered similar to us are \cite{li2009contour,zhang2010smarter}. Li et al \cite{li2009contour} proposed a two-layer tracking framework in which PSO is successfully combined with a level set evolution. In the first layer, PSO is adopted to capture the global motion of the target and to help construct the coarse contour. In the second layer, level set evolution based on the coarse contour is carried out to track the local deformation. However, there is a possibility that most samples will get trapped in a few strong local maxima. Hence, the PSO method fails to track highly abrupt motions. Zhang et al. \cite{zhang2010smarter} proposed a swarm intelligence based particle ﬁlter algorithm with a hierarchical importance sampling process which is guided by the swarm intelligence extracted from the particle conﬁguration, and thus greatly overcome the sample impoverishment problem suffered by particle ﬁlters. Unfortunately, it cannot be a perfect solution either as it is still depends on the Gaussian approximation. In order to handle this issue, we introduce 1) DAP - a dynamic acceleration parameters by utilising the averaged velocity information of the particles; and 2) $\mathcal{EF}$ - a mechanism that balance the tradeoffs between exploration and exploitation of the swarm into the PSO framework. With this, the SwaTrack will able to alleviate these problems and track in highly abrupt motion or recover from tracking error.

\section{Particle Swarm Optimisation Revisit}
\label{sec:ParticleSwarm}

Particle Swarm Optimisation (PSO) - a population-based stochastic optimisation technique was developed by Kennedy and Eberhart in 1995 \cite{Eberhart}. It was inspired by the social behaviour of a flock of birds. The operation of the PSO can be described as let us assume an $n$-dimensional search space, $\mathcal{S}$ $\subset$ $\mathcal{R}^n$ and a swarm comprising of $I$ particles. Each particle represents a candidate solution to the search problem and is associated to a fitness function (cost function), $f:\mathcal{S}\to \mathcal{R}$. At every $k$th iteration, each particle is represented as ${\{x_k^i\}_{i=1,...,I}}$ at $k$th iteration, where $k=1,2,...K$. Each particle, $x_k^i$ has its own velocity, $v(x_k^i)$ and a corresponding fitness value (cost), $f(x_k^i)$. The best position encountered by the $i$th particle (personal best) will be denoted as ${\{p(x_k^i)\}_{i=1,...,I}}$ and the fitness value as $pBest_k^i=f(p(x_k^i))$. For every $k$th iteration, the particle with the best fitness value will be chosen as the global best and is denoted as the index of the particle is denoted as $f$. Finally, the overall best position found by the swarm will denote as $gBest_k^i=f(p(x_k^g))$. The PSO algorithm is shown in Algo 1:

\begin{enumerate}
\item \textbf{Initialisation}, at iteration $k=0$
\begin{itemize}
\item Initialise a population of $I$ particles, ${\{x_k^i\}_{i=1,...,I}}$ with positions, $p(x_k^i)$,at random within the search space, $\mathcal{S}$.
\item Initialise the velocities, $v(x_k^i)$ at random within $[1,-1]$.
\item Evaluate the fitness value of each particle and identify their personal best $pBest_k^i=f(p(x_k^i))$.
\item Identify the global best $g$th particle and update the global best information, $gBest_k=f(p(x_k^g))$. 
\end{itemize}

\item \textbf{Repeat} at iteration $k=1,2,...K$ until the stopping criterion is met.
\begin{itemize}
\item For each $i$th particle, compute the new velocity according to:
\begin{equation}\label{eq:velocity}
\begin{split}
v_{k+1}^i = [(\omega*v_k^i) + (c_1*r_1*(pBest_k^i-x_k^i)) 
+ (c_2*r_2*(gBest_k-x_k^i)]
\end{split}
\end{equation}
\item For each $i$th particle, move them using the computed new velocity as in Eq. \ref{eq:velocity} and update its position according to:
\begin{equation}\label{eq:update}
p(x_{k+1}^i) = p(x_k) ^i+ v_{k+1}^i 
\end{equation}
\item For each $i$th particle, ensure the newly computed position is within state space, $p(x_{k+1}^i) \subset \mathcal{S}$ 
\item Update $pBest_k^i$, $p(pBest_k^i)$, $g$, $gBest_k$, $p(gBest_k^g)$.
\item Check for \textbf{Convergence}
\item \textbf{End Repeat}
\end{itemize}
\end{enumerate}

The parameters $\omega$, $c_1$ and $c_2$ are positive acceleration constants used to scale the influence of the inertia, cognitive and social components respectively; $r_1$, $r_2$ $\subset$ $(0,1)$ are uniformly distributed random numbers to randomise the search exploration. 

\subsection{Limitations for Abrupt Motion Tracking}

However, the traditional PSO does not able to cope with abrupt motion, due to few reasons: 

{\bf Constant Acceleration Parameters:} The parameter $c_1$ controls the influence of the cognitive component, $(c_1*r_1*(pBest_k^i-x_k^i))$ which represents the individual memory of particles (personal best solution). A higher values gives more emphasize on the cognitive component and vice versa. In contrast, the parameter $c_2$ controls the influence of the social component,$(c_2*r_2*(gBest_k-x_k^i)$ which indicates the joint effort of all particles to optimize a particular fitness function, $f$. 

A drawback of the current PSO is the lack of a reasonable mechanism to effectively handle the acceleration parameters ($\omega$, $c$ and $r$); which always set to a constant variable. For example, many applications of the PSO and its variant set these parameters $c_1 = c_2 = 2.00$, which gives the stochastic factor a mean of 1.0 and giving equal importance to both the cognitive and social components \cite{Eberhart}. This limits the search space and therefore could not cope with abrupt motion. Therefore, it is essential to have dynamic acceleration parameters that able to cope with unexpected dynamic motion.

{\bf Tradeoffs in Exploration and Exploitation:} The inertia weight, $\omega$ serves an important value that directs the exploratory behaviour of the swarms. High inertia weights accentuates the influence of previous velocity information and force the swarm to explore a wider search space; while a decreasing inertia weight reduces the influence of previous velocity and exploit a smaller search space. Often, the inertia value that controls the influence of the previous velocity is set to $\omega \in [0.8,1.2]$ \cite{Shi}. Recently, decaying inertia weigh, $\omega = 0.9 \to 0.4$ has been proposed and tested, with the aim of favouring global search at the start of the algorithm and local search later. While these settings has been tested to work well in other optimisation problems, one must note that it is not applicable in tracking abrupt motion where the dynamic change is unknown. Therefore, a solution that able to handle the tradeoffs between the exploration and exploitation will be beneficial.



\section{Proposed Method - SwaTrack}
\label{sec:proposedAPSO}

In this section, we present our proposed SwaTrack - a variant of the traditional PSO to track target with arbitrary motion. Particularly, we will discuss how the ingenious combination of Dynamic Acceleration Parameters (DAP) and Exploration Factor $\mathcal{EF}$ in the proposed PSO framework has alleviated the problem of swarm explosion and divergence problem.

\subsection{Dynamic Acceleration Parameters (DAP)}

PSO is a population based stochastic optimization technique. Since PSO is an iterative solution, efficient convergence is an important issue toward a real time abrupt motion estimation system. However, the strict threshold of the conventional PSO velocity computation as in Eq. \ref{eq:velocity} will always lead to particles converging to a common state estimate (the global solution). One reason is that the velocity update equation uses a decreasing inertia value which indirectly forces the exploration of particles to decrease over the iterations. On the other hand, an increasing inertia value will lead to swarm explosion in some scenarios. 

To overcome this, we introduce DAP - a mechanism to self-tune the acceleration parameters by utilising the averaged velocity information of the particles. That is, we normalised the acceleration parameters so that they can be compared fairly with respect to the estimated velocity, ${p(w \cap c_1 \cap c_2) = 1}$. The fitness function information is incorporated in the PSO framework in order to refine the acceleration parameters dynamically rather than a static value. The basic idea is that when an object moves consistently in a particular direction, $\mathcal{C} \to 1$, the inertia, \emph{w} and cognitive weight, \emph{c${_1}$} values are increased to allow resistance of any changes in its state of motion in the later frames. Otherwise when $\mathcal{C} \to 0$, the social weight \emph{c${_2}$} is increased by a stepsize to reduce its resistance to motion changes as Eq. \ref{eq1q}. The increase of the social weight allows global influence and exploration of the search space, which is relevant when the motion of a target is dynamic. The exploitation within nearby regions is reasonable when an object is moving with small motion. 

\begin{equation}
\label{eq1q}
\left\{ \begin{array}{l}
{c_1} = {c_1} + m;\,\, {\rm{ }}{c_2} = {c_2} - m;\,\,\ \omega = \omega + m{\rm{ }}\,\,\,\,\,\,\,\,\,\,\,\,\,C \to 1\\ \\
{c_1} = {c_1} - m;\,\,{\rm{ }}{c_2} = {c_2} + m;\,\, \omega = \omega - m{\rm{ }}\,\,\,\,\,\,\,\,\,\,otherwise\\ \\ 
{\rm{^*condition \, on }}\,\,\,p(\omega \cap {c_1} \cap {c_2}) = 1
\end{array} \right.
\end{equation}

The $C$ is estimated by computing the frequency of the change in the quantised motion direction of the object; $\mathcal{C} \to 1$ represent consistent motion with minimal change of direction, while $\mathcal{C} \to 0$ represent inconsistent or dynamic motion. 


\subsubsection{Exploration Factor ($\mathcal{EF}$)}

The normalisation of DAP as $1$ will restrict the overall exploration of the state to a certain degree. Hence, we refine this by introducing the exploration factor, $\mathcal{EF}$ which serves as a multiplying factor to increase or decrease the exploration. We define the exploration factor, $\mathcal{EF}$ as the parameters that adaptively
\begin{enumerate}
\item increase the $exploration$ with high variance, and 
\item increase the $exploitation$ with low variance. 
\end{enumerate}

By utilising these exploitation and exploration abilities, our method is capable to recover from being trapped into a common state (local optima). Thus, the proposed SwaTrack copes better with both the smooth and abrupt motion. At every $k$th iteration, the quality of the estimated position upon convergence (global best) is evaluated using its fitness value. $f(gBest_k^g) \to 1$ indicates high likelihood whereas $f(gBest_k^g) \to 0$ indicates low likelihood or no similarity between an estimation and target.

When $f(gBest_k^g) \le T_{MinF}$, where $T_{MinF}$ is a threshold, we know that there is low resemblance between the estimation and target and most likely the proposal distribution may not match the actual posterior. Thus in this scenario, $\mathcal{EF}$ is increased alongside the maximum number of iterations, $K$ by empirically determined step sizes \emph{m and n} respectively. This drives the swarm of particles to explore the region beyond the current local maxima (increase exploration). However, when an object has left the scene, $K$ tend to increase continuously and cause swarm explosion. Thus, we limit $K \subset \mathcal{S}$.

\begin{equation}\label{eq:swatrackexploration}
\mathcal{E} \;\alpha \; f(gBest_k^g)
\end{equation}

In another scenario, where $f(gBest_k^g) \ge T_{MinF}$, $\mathcal{EF}$ is decreased alongside \emph{K}; constraining the search around the current local maximum (exploitation). In a straightforward manner, it is always best to drive particles at its maximum velocity to provide a reasonable bound in order to cope with the maximum motion change. However, this is not reasonable for real-time applications as it incurs unnecessary computational cost especially when the motion is not abrupt. Thus, by introducing the adaptive scheme to automatically adjust the exploration and exploitation behaviour of the swarm, SwaTrack is able to cope with both the smooth and abrupt motion with less computational cost. Also, we observed that since the particles in SwaTrack exchange information with one another, we only need a few particles for sampling.



\subsection{Novel Velocity Model}

With the introduction of DAP and $\mathcal{EF}$, the novel velocity model, $v'$ in our PSO framework is represented as:

\begin{equation}\label{eq:swatracktwo}
\begin{split}
\grave v_{k+1}^i = \mathcal{EF}_k[(\omega*\grave v_k^i) + (c_1*r_1*(pBest_k^i-x_k^i))
+ (c_2*r_2*(gBest_k-x_k^i)]
\end{split}
\end{equation}

\noindent where $\mathcal{EF} _k$ is the exploration factor at iteration $k$ and $c,\: r,\:\omega$ are the acceleration parameters with the condition ${p(w \cap c_1 \cap c_2) = 1}$. The normalised condition applied to the acceleration allows on the fly tuning of these parameters according to the quality of the fitness function. The fitness function used here is represented by the normalised distant measure between the appearance model of an estimation and the object-of-interest. The fitness value of a particle, $f(x_k^i)$ measures how well an estimation of the object's position matches the actual object-of-interest; where $1$ represents the highest similarity between an estimation and target and $0$ represents no similarity.

At every $k$th iteration, each particle varies its velocity according to Eq. \ref{eq:swatracktwo} and move its position in the search space according to:

\begin{equation}\label{eq:swatrackupdate}
p(x_{k+1}^i) = p(x_k) ^i+ \grave{v}_{k+1}^i 
\end{equation}

Note that the motion of each particle is directed towards the promising region found by the global best, $gBest_k$ from previous iteration, $k=k-1$. 



\begin{enumerate}
\item \textbf{Initialisation}, at iteration $k=0$
\begin{itemize}
\item Initialise a population of $I$ particles, ${\{x_k^i\}_{i=1,...,I}}$ with positions, $p(x_k^i)$,at random within the search space, $\mathcal{S}$.
\item Initialise the velocities, $v(x_k^i)$ at random within $[1,-1]$.
\item Evaluate the fitness value of each particle and identify their personal best $pBest_k^i=f(p(x_k^i))$.
\item Identify the global best $g$th particle and update the global best information, $gBest_k=f(p(x_k^g))$. 
\end{itemize}

\item \textbf{Repeat} at iteration $k=1,2,...K$ until the stopping criterion is met.
\begin{itemize}
\item For each $i$th particle, compute the new velocity according to:
\begin{equation*}\label{eq:velocity}
\begin{split}
\grave v_{k+1}^i = \mathcal{E}_k[(\omega*\grave v_k^i) + (c_1*r_1*(pBest_k^i-x_k^i))
+ (c_2*r_2*(gBest_k-x_k^i)] \\
\end{split}
\end{equation*}
where, 
\begin{equation*}
p(w \cap c_1 \cap c_2) = 1
\end{equation*}
\item \textbf{If} 
\;\; $f(gBest_k^g) \le T_{MinF}$ \\
\; \textbf{then}
\;\; $\mathcal{E}=\mathcal{E}+n$, $K=K+n$ \\
\; \textbf{else} \\
\; \textbf{then}
\;\; $\mathcal{E}=\mathcal{E}-n$, $K=K-n$ \\
\textbf{end If}
\item \textbf{If} 
\;\; $\mathcal{C} \to 1$\\
\; \textbf{then}
\;\; $c_1=c_1+m$, $c_2=c_2-m$, $\omega=\omega+m$ \\
\; \textbf{else} \\
\; \textbf{then}
\;\; $c_1=c_1-m$, $c_2=c_2+m$, $\omega=\omega-m$ \\
\textbf{end If}
\item For each $i$th particle, move them using the computed new velocity as in Eq. \ref{eq:velocity} and update its position according to:
\begin{equation*}\label{eq:update}
p(x_{k+1}^i) = p(x_k) ^i+ \grave{v}_{k+1}^i 
\end{equation*}
\item For each $i$th particle, ensure the newly computed position is within state space, $p(x_{k+1}^i) \subset \mathcal{S}$ 
\item Update $pBest_k^i$, $p(pBest_k^i)$, $g$, $gBest_k$, $p(gBest_k^g)$.
\item Check for \textbf{Convergence}
\item \textbf{End Repeat}
\end{itemize}
\end{enumerate}

\section{Experimental Results \& Discussion}
\label{sec:experiment}

In this section, we verify the feasibility and robustness of our proposed method in handling abrupt motion via various experiments using public and synthetic datasets. The experiments were performed on an Intel Core-2 processor with C++ and OpenCV implementation.

\subsection{Experimental Settings}

We assumed the object-of-interest to be known and hence initialise manually the 2D position of the target in the first frame as automatic initialisation is another research topic by itself. The object is represented by its appearance model, which comprises of HSV histogram with uniform binning; 32 bins. The normalised Bhattacharyya distant measure is used as the fitness value (cost function) to measure the quality of the estimation; where 1 represents the highest similarity between an estimation and target and 0 represents no similarity. Here, the initial values for SwaTrack are $\mathcal{EF}=25$, $\omega=0.4$, $c_1=0.3$, $c_2=0.3$, $K=30$, $I=15$ respectively. These values are set empirically and are not as critical; the adaptive mechanism in the proposed method allows adjustment of these parameters according to the quality of the observation model. 

We manually labelled the ground truth of $n$th object. The ground truth is described as bounding box information, \emph{X$^{n}$(x$^{n}$,y$^{n}$,w$^{n}$,h$^{n}$ = x-positions, y-positions, width, height)}. We compare the state-of-the-art results of PSO, PF \cite{Yan,Maggio}, BDM \cite{Wong}, FragTrack \cite{Adam}, A-WLMC \cite{Kwon2008} and CT \cite{zhang2008sequential}, respectively in terms of both the detection accuracy (\%) and processing time (milliseconds per frame).


\begin{figure}[htb]
\begin{center} 
\includegraphics[height=0.35\linewidth, width=0.85\linewidth]{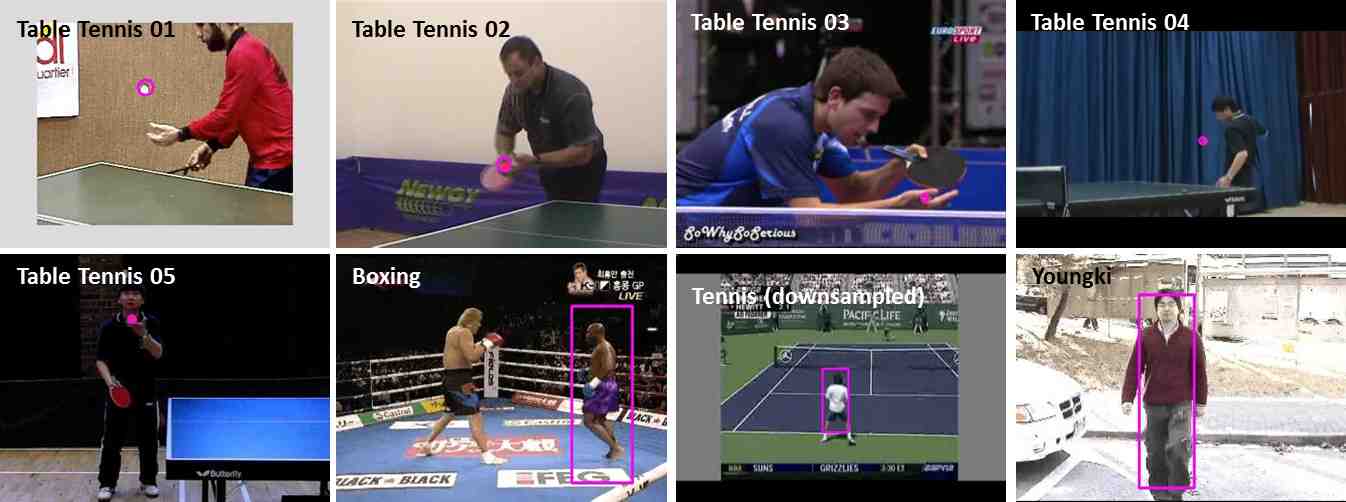}
\caption{Sample shots of the dataset employed.} 
\label{fig:sampledata01} 
\end{center}
\end{figure}

\subsection{Dataset}

We tested our proposed method with a number of public and synthetic datasets. In the quantitative experiments, we employed 5 public datasets (TableTennis, Youngki, Boxing and Tennis) as illustrated in Fig. \ref{fig:sampledata01}. {\bf Rapid Motion of Small Object}: The $TableTennis (TableT)$ dataset consists of 5 video sequences to test the effectiveness of the proposed method in terms of tracking small object (e.g. table tennis ball) that exhibits fast motion. Sequence 1\footnote {The SIF training data is available at http://www.sfu.ca/~ibajic/datasets.html}, the SIF Table Tennis sequence is a widely used dataset in the area of computer vision, especially for evaluation of detection and tracking methods. This sequence has complex, highly textured background and exhibit camera movement with some occlusion between the ball and the player's arm. Sequence 2\footnote {The ITTF training and demo dataset is available at http://www.ittf.com}, is a sample training video from the ITTF video library which is created to expose players, coaches and umpires to issues related to service action. Although this sequence is positioned to provide the umpire's point of view of a service, it is very challenging as the size of the tennis ball is very small; about 8x8 pixels to 15x15pixels for an image resolution of 352x240. The video comprises of 90 frames including 10 frames in which severe occlusion happens, where the ball is hidden by the player's arm. Sequence 3\footnote{The video is available at http://www.youtube.com/watch?v=C9D88AcmLjI} is a match obtained from a publicly available source. In this sequence, the tennis ball is relatively large as it features a close-up view of the player. However, there are several frames where the ball appears to be blurred due to the low frame rate and abrupt motion of the tennis ball. Sequence 4 and 5\footnote{This dataset can be obtained from http://xgmt.open.ac.uk} is captured at a higher frame rate, thus the spatial displacement of the ball from one frame to another appears to be smaller (less abrupt) and the ball is clearer. This is to test the ability of proposed method to handle normal visual tracking scenario. Since the dataset didnt provide any groundtruth, we manually labelled the ground truth of $n$th object. The ground truth is described as bounding box information, \emph{X$^{n}$(x$^{n}$,y$^{n}$,w$^{n}$,h$^{n}$ = positions in the x-dimension, y-dimension, width, height)}. If the paper is accepted, we will make the dataset and groundtruth publicly available. {\bf Switching Camera}: The $Youngki$ and $Boxing$ dataset \cite{Kwon2012} comprise of frames edited from different cameras (camera switching). This results in an object appearing at the different part of the image. {\bf Low-frame rate} $Tennis$ dataset \cite{Kwon2012} comprises of downsampled data to simulate abrupt change. The frames are downsampled from a video with more than 700 original frames, by keeping one frame in every 25frames. The rapid motion of the tennis player from one frame to another due to the downsampling made tracking extremely difficult. Downsampling is done to simulate abrupt motion during low-frame rates. 

In the qualitative experiments, on top of the 5 public datasets, we included 1 more dataset that consists of 3 video sequences to further test the robustness of the proposed method. {\bf Abrupt Motion with Inconsistent Speed:} The first sequence of the dataset is to track a synthetic ball which moves randomly across the sequence with inconsistent speed, whilst the second sequence tracks a soccer ball which is being juggled in a free-style manner in a moving scene with a highly textured background (grass). {\bf Multiple targets:} In the third sequence is two simulated balls moving at random. We intend to demonstrate the capability of the proposed system to track multiple targets; whilst most of the existing solutions are focused on single target.


\subsection{Quantitative Results}

\subsubsection{Experiment 1: Detection Rate}

Detection rate refers to the correct number and placement of the objects in the scene. For this purpose, we denote the ground truth of \emph{n}th object as \emph{GT$_n$} and the output from the tracking algorithms of \emph{j}th object is denoted as \emph{$\xi_n$}. We describe the ground truth and tracker output of each \emph{n}th object as bounding box information , \emph{X$^{n}$(x$^{n}$,y$^{n}$,w$^{n}$,h$^{n}$ = x-position, y-position, width, height)}. The coverage metric determines if a \emph{GT} is being tracked, or if an \emph{$\xi$} is tracking accurately. In \cite{Smith}, it is shown that the \emph{F-measure}, $F$ suited this task as the measure is 1.0 when the estimate, $\xi_n$ overlaps perfectly with the ground truth, \emph{GT$_n$}. Two fundamental measures known as \emph{precision} and \emph{recall} are used to determine the \emph{F-measure}. 

\emph{Recall}: Recall measures how much of the \emph{GT} is covered by the \emph{$\xi$} and takes value of 0 if there is no overlap and 1 if they are fully overlapped. Given a ground truth, \emph{GT$_n$} and a tracking estimate, \emph{$\xi_n$}, the \emph{recall}, \emph{$\Re_n$} is expressed as:

\begin{equation}\label{eq:recall}
\Re_n = \dfrac{|\xi _n \cap GT_n|}{|GT _n|}
\end{equation}

\emph {Precision}: Precision measures how much of the \emph{$\xi$} covers the \emph{GT} takes value of 0 if there is no overlap and 1 if they are fully overlapped. The \emph{precision}, \emph{$\wp_n$} is expressed as:

\begin{equation}\label{eq:recall}
\wp_n = \dfrac{|\xi _n \cap GT_j|}{|\xi _n|}
\end{equation}

\emph {F-measure}: The \emph{F-measure}, \emph{F$_n$} is expressed as:
\begin{equation}\label{eq:fmeasure}
F_n = \dfrac{2\Re_n \wp_n}{\Re_n+\wp_n}
\end{equation}

{\bf Coverage Test} In this experiment, we employ the $F-measure$ according to the score measurement of the known PASCAL challenge \cite{Everingham}. That is, if the \emph{F$_n$} of \emph{n}th object is larger than 0.5, the estimation is considered as correctly tracked in the frame. Table. \ref{overalldetectionaccuracy} demonstrates the detection accuracy of the benchmarked tracking algorithms for all 8 test sequences. Overall, the experimental results show that the average tracking accuracy of the proposed method surpasses most of the state-of-the art tracking methods with an average detection accuracy of 91.39\%. For all 6 test sequences (TableT1, TableT2, TableT5, $Youngki$ and $Tennis$), the SwaTrack generates the best tracking results amongst the rest and Rank 2 for TableT4 and $Boxing$, respectively. 

\begin{figure}[htbp]
\centering
\subfigure[Sample detections from PF tracking.]{\includegraphics[height=0.25\linewidth, width=0.85\linewidth]{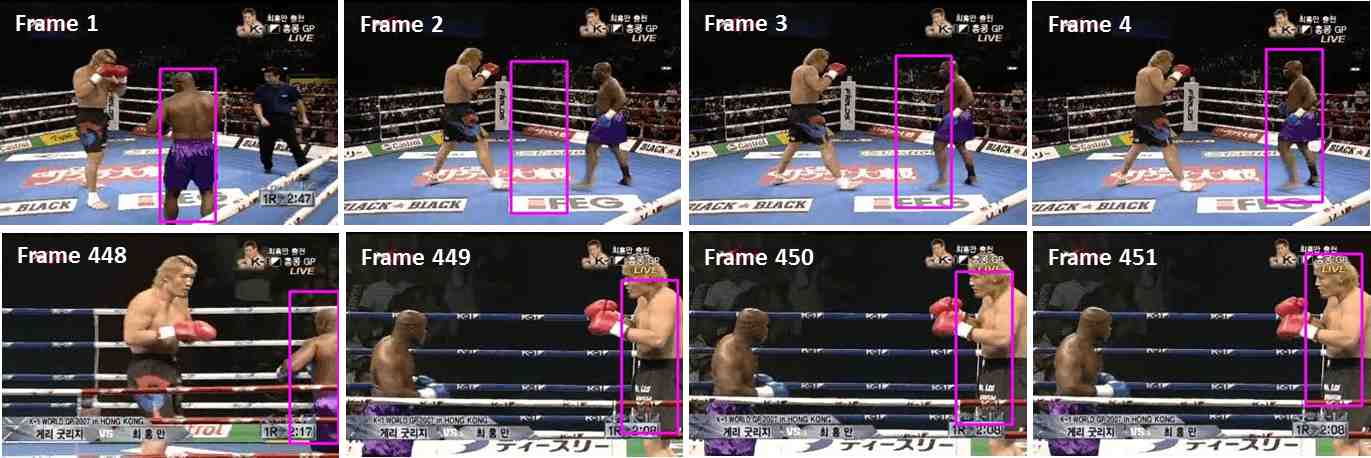}
\label{fig:LO1}}
\subfigure[Sample detections from SwaTrack tracking. ]{\includegraphics[height=0.25\linewidth, width=0.85\linewidth]{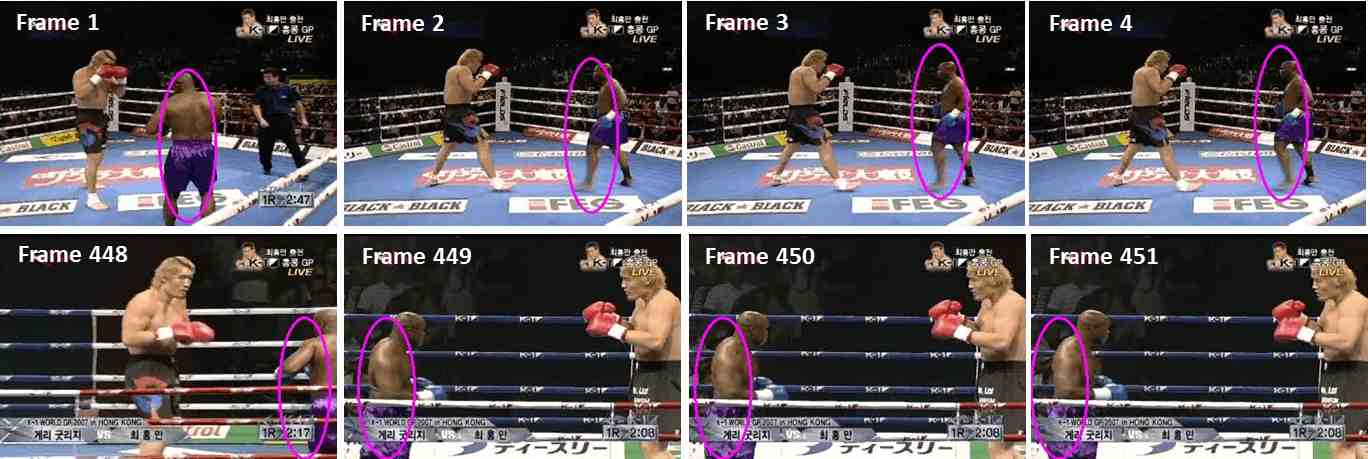}
\label{fig:LO2}}
\caption{Sample output to demonstrates incorrect tracking due to trapped in local optima. The aim is to track the person in dark skin and purple short. From Frame 449-451 (a), PF lost track of the object due to sampling from incorrect distribution during abrupt motion.. On the other hand, the results in (b) demonstrate the capability of the SwaTrack tracking in dealing with the non-linear and non-Gaussian motion of the object.}
\label{fig:LocalOptimaPFAPSO}
\end{figure}

\begin{figure}[h]
\centering
\subfigure[Sample detections from PF.]{\includegraphics[height=0.2\linewidth, width=0.8\linewidth]{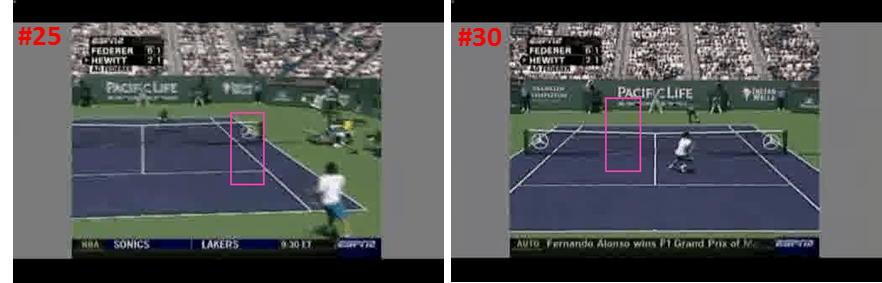}
\label{fig:LO1a}}
\subfigure[Sample detections from SwaTrack.]{\includegraphics[height=0.2\linewidth, width=0.8\linewidth]{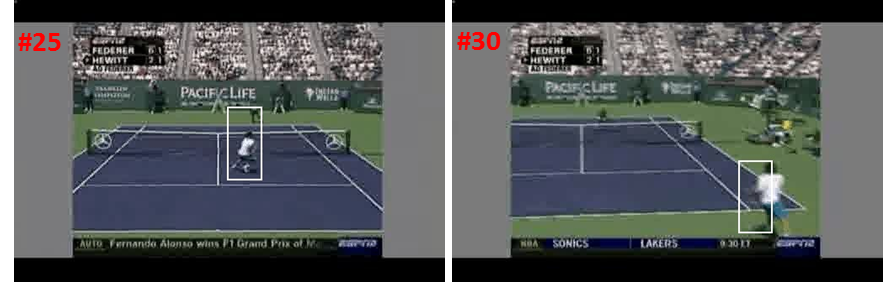}
\label{fig:LO2a}}
\subfigure[Sample detections from A-WLMC.]{\includegraphics[height=0.2\linewidth, width=0.8\linewidth]{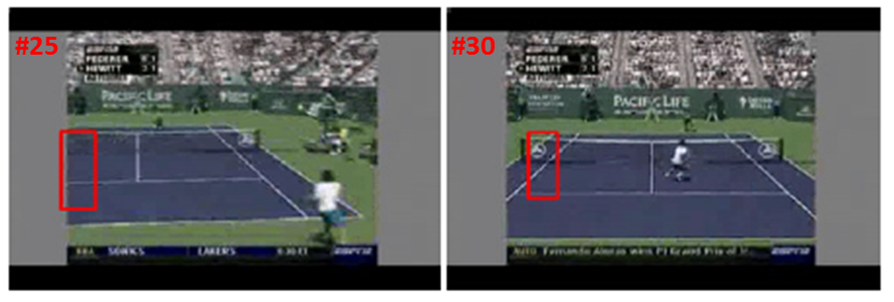}
\label{fig:LO1a}}
\subfigure[Sample detections from IA-MCMC.]{\includegraphics[height=0.2\linewidth, width=0.8\linewidth]{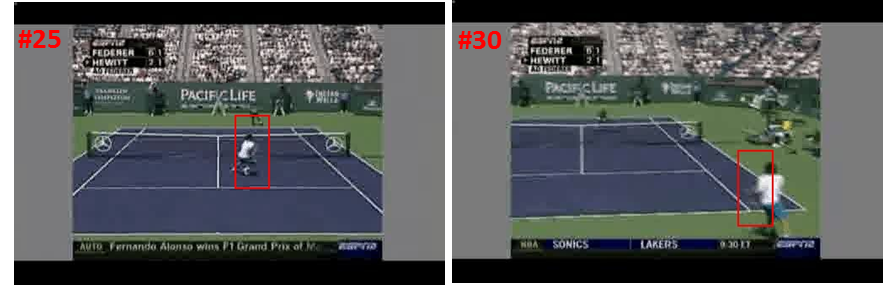}
\label{fig:LO2a}}
\caption{A comparison between PF, SwaTrack, A-WLCM \cite{Kwon2012} and IA-MCMC \cite{Zhou}. It is observed that the SwaTrack tracking gives a more accurate fit of the state. }
\label{fig:LocalOptimaMCMCAPSO}
\end{figure}

In the meantime, methods that do not built on sophisticated motion model, the FragTrack \cite{Adam} employs refine appearance model that adapts to the changes of the object. Even so, it still performs poorly in this condition when compared to the others with an average accuracy of 37.19\%. PF on the other hand, gives the detection accuracy of 85.6\%. This is expected as it is known that PF algorithm is constraint to a fixed Gaussian motion model. Once PF has lost track of the object, it has the tendency to continue searching for the object in the wrong region such as shown in Fig. \ref{fig:LocalOptimaPFAPSO}a; leading to error propagation and inability to recover from incorrect tracking. The proposed method copes better with abrupt motion and is not subjected to trapped in local optima as shown in Fig. \ref{fig:LocalOptimaPFAPSO}b. While the MCMC tracking method is still subjected to a certain degree of recovery as shown in Fig. \ref{fig:LocalOptimaMCMCAPSO}. 

\begin{table}[t]
\caption{Experiment results - Comparison of the Detection Rate (in $\%$)}
\label{overalldetectionaccuracy}
\centering
\begin{tabular}{|c||c|c|c|c|c|c|c|}
\hline \hline
&PSO & PF \cite{Yan,Maggio} & BDM \cite{Wong} & FragTrack \cite{Adam} & A-WLMC \cite{Kwon2008} & CT \cite{zhang2008sequential} & SwaTrack\\
\hline
\hline
TableT1 & 70.1 & 58.4 & 68.3 & 64.9 & 47.2 & 72.3 & {\bf 87.8} \\
\hline
TableT2 & 83.1 & 69.8 & 53.4 & 24.1 & 3.2 & 4.3 & {\bf 93.1} \\
\hline
TableT3 & 58.2 & 52.1 & 67.3 & 55.3 & 8.7 & 24.5 & {\bf 74.1} \\
\hline
TableT4 & 59.6 & 47.3 & 73.2 & 57.2 & 6.9 & {\bf 98.2} & 97.3 \\
\hline
TableT5 & 60.3 & 34.5 & 64.2 & 9.7 & 5.4 & 36.3 & {\bf 72.8} \\
\hline \hline
Average & 66.26 & 52.42 & 65.28 & 42.24 & 14.28 & 47.12 & {\bf 85.02} \\
\hline \hline
\end{tabular}
\end{table}

\begin{table}[ht]
\caption{Experiment results - Comparison of the Detection Rate (in $\%$)}
\label{overalldetectionaccuracy1}
\centering
\begin{tabular}{|c||c|c|c|c|c|c|}
\hline \hline
&PSO & PF \cite{Yan,Maggio} & FragTrack \cite{Adam} & A-WLMC \cite{Kwon2008} & SwaTrack\\
\hline
\hline
Tennis & 87.3 & 67.3 & 20.6 & 95.1 & 98.3 \\
\hline
Youngki & 87.1 & 47.2 & 27.5 & 86.8 & 98.7 \\
\hline
Boxing & 82.4 & 16.3 & 48.3 & 98.1 & 96.3 \\
\hline \hline
Average & 85.6 & 43.6 & 32.13 & 93.33 & {\bf 97.76} \\
\hline \hline
\end{tabular}
\end{table}

{\bf Dataset Unbias} The problem of dataset bias was highlighted in \cite{torralba2011} where the paper argue that ``Is it to be expected that when training on one dataset and testing on another there is a big drop in performance?'' In here, we replicate similar scenario in tracking domain and observe that though the A-WLMC method \cite{Kwon2008} performs well in TableT4 and $Youngki$ sequence, they do not produce consistent results when tested across the other datasets as shown in Table \ref{overalldetectionaccuracy}-\ref{overalldetectionaccuracy1} . For example, we can notice that the average detection accuracy of A-WLMC is 14.28\% for TableT dataset, and 93.33\% for Tennis, Boxing and Youngki dataset respectively. The indicate that the A-WLMC solution \cite{Kwon2008} is dataset bias as it seem to only work well in their proposed dataset, but performed porrly when is employed on different dataset. Perhaps this is due to the motion model employed by these tracking methods that works well only on certain scenarios, alluding to the notion in \cite{Garcia} that different motion requires different motion models. This is indeed not the case for our proposed SwaTrack. Our overall detection rates are 85.02\% and 97.76\%, respectively. For all sequences that exhibit different challenging conditions, e.g. rapid motion ($TableT1-5$), low-frame rate ($Tennis$), the Swatrack has shown its robustness to cope without any ease. 

We further investigated the dataset bias problem and found out that there is an influence of object size to the detection rate. For instance, A-WLMC algorithm \cite{Kwon2008} performs poorly for sequences in which the resolution of the object-of-interest is small, such as in Table Tennis dataset and performs surprisingly well when the object is large such as in the $Youngki$, $Boxing$ and $Tennis$ dataset, respectively. This indicates the need to have better representation of the object for a more accurate acceptance and rejection of estimations in the MCMC. 

\begin{figure}[htbp]
\centering 
\includegraphics[width=\textwidth, height=6cm]{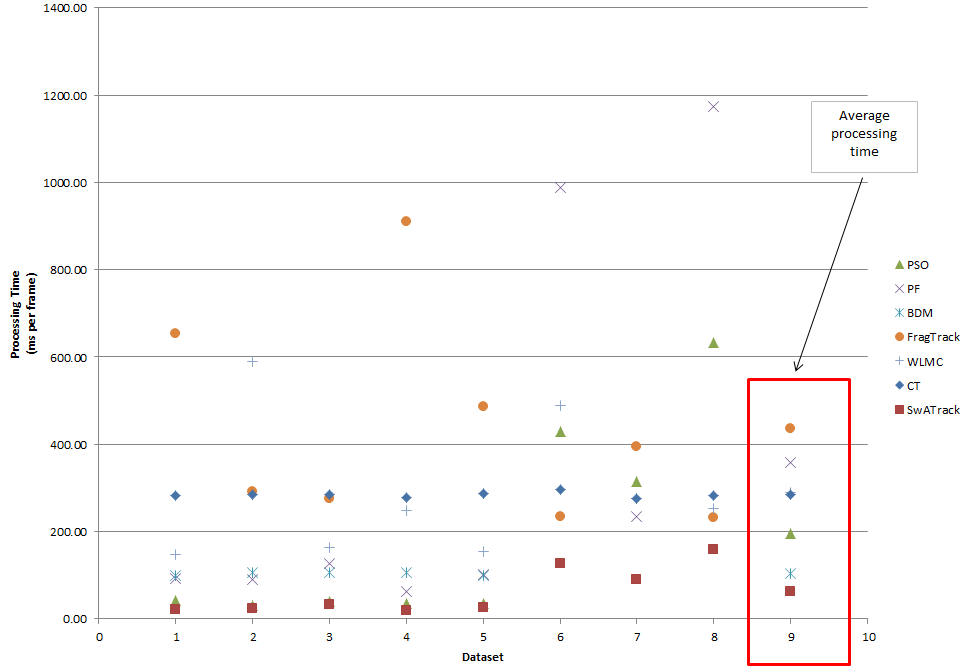}
\caption{Time Complexity. This figure illustrates the comparison of processing time (milliseconds per frame) between the proposed SwaTrack, standard PSO, PF, BDM, FragTrack, A-WLMC and CT}
\label{fig:overallprocessingtime}
\end{figure}

\subsubsection{Experiment 2: Computational Cost}

Fig. \ref{fig:overallprocessingtime} demonstrates the comparison of the proposed method with state-of-the-art solution in terms of time complexity. As for the processing time, the SwaTrack algorithm requires the least processing time with an average of 63 milliseconds per frame. In contrary, MCMC-based solution such as A-WLMC \cite{Kwon2008} and PF \cite{Yan,Maggio} require higher processing time. This is likely due to the inherent correlation between MCMC samplers which suffer from slow convergence when an object has not been tracked accurately. Notice that in scenarios where the MCMC requires high processing time, the accuracy of the MCMC is minimal; the increase in computational cost is due to the increase of search space when the observation model is unlikely representing the target. Note that the optimal number of samples deployed in the PF and MCMC throughout the sequences has been selected empirically; where it ranges from 150 to 1000 particles in PF, 600 to 1000 particles in MCMC with 600 iterations while SwaTrack uses 10-50 particles with 5-70 iterations. Intuitively, an increase in the number of samples would lead to an increase in computational cost as each particle would need to be evaluated against the appearance observation; elucidating the minimal processing time required by the proposed SwaTrack.

As shown in Table \ref{overalldetectionaccuracy}-\ref{overalldetectionaccuracy1}, in which the SwaTrack detection rate is Rank 2, though the CT \cite{zhang2008sequential} and A-WLMC \cite{Kwon2008} achieved better but their average processing time are almost increase by 3x compared to SwaTrack. As compared to the A-WLMC tracking which increases its subregions for sampling when the state space increases, our method adaptively increases and decreases its proposal variance for a more effective use of samples. Thus the processing time required is much lesser as compared to the other methods. The advantage of the dynamic mechanism is shown when comparing the processing time of SwaTrack to PSO (average of 195.20 milliseconds per frame); where the processing time of PSO is 3x more than that of SwaTrack. In summary, the experimental results demonstrate the capability of the proposed system to cope with the variety of scenario that exhibits highly abrupt motion. The adaptation of a stochastic optimisation method into tracking abrupt motion has been observed to incur not much additional processing cost, yet at the same time is able to have fair tracking accuracy as compared to the more sophisticated methods. Thus, the preliminary results give a promising indication that sophisticated tracking methods may not be necessary after all.

\subsection{Qualitative Results}

%

{\bf Low Frame Rate} The sequence aims to track a tennis player in a low-frame rate video, which is down-sampled from a 700 frames sequence by keeping one frame in every 20 frames. Here, the target (player) exhibits frequent abrupt changes which violate the smooth motion and constant velocity assumptions. Thus, motion that is governed by Gaussian distribution based on the Brownian or constant-velocity motion models will not work in this case. Fig. \ref{fig:LocalOptimaMCMCAPSO} shows sample shots to compare the performance between conventional PF tracking (500 samples), A-WLMC (600 samples) \cite{Kwon2012}, IA-MCMC (300 samples)\cite{Zhou} and SwaTrack (50 samples). It is observed that the tracking accuracy of SwaTrack is better than PF and A-WLMC even by using fewer samples. While the performance of SwaTrack is comparable to IA-MCMC, SwaTrack requires fewer samples and thus requires less processing requirement. These results further verify that the proposed method is able to track the moving targets accurately and effectively, regardless of the variety of change in the target's motion.

\begin{figure}[htbp]
\centering
\subfigure[Sample of SwaTrack on $Boxing$ sequence. ]{\includegraphics[height=0.15\linewidth, width=0.8\linewidth]{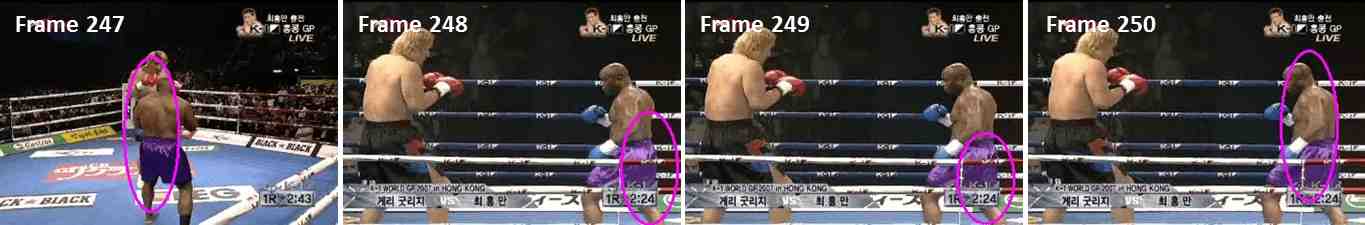}
\label{fig:LO3a}}
\subfigure[Sample of SwaTrack on $Youngki$ sequence. ]{\includegraphics[height=0.15\linewidth, width=0.8\linewidth]{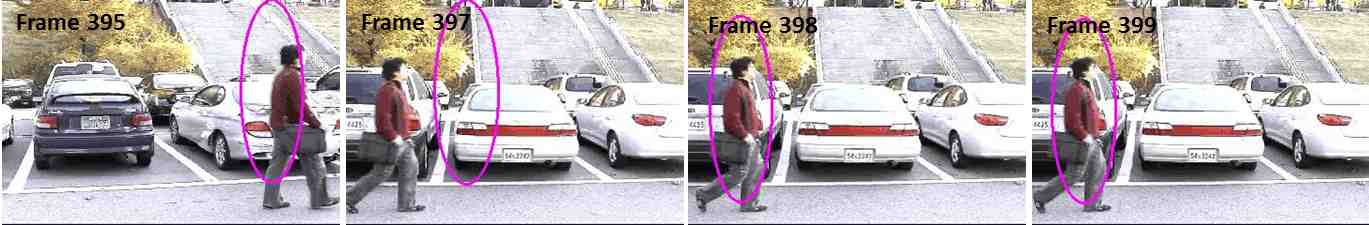}
\label{fig:LO4a}}
\caption{Sample outputs to demonstrate the flexibility of the proposed SwaTrack to recover from incorrect tracking. Minimal frames (1-2frames) are required to escape from local optima and achieve global maximum.}
\label{fig:Recovery}
\end{figure}

{\bf Local Miminum Problem} In this experiment, we aim to test the capability of SwaTrack to recover from incorrect tracking. This is to test the capability of the DAP and $\mathcal{EF}$ to handle the abrupt motion. Fig. \ref{fig:Recovery} shows the result for Youngki where the camera switches. Due to this phenomenon, the subject will have a drastic change of position the adjacent image. It can be seen that the SwaTrack is able to cope with this problem. 

\begin{figure}[htbp]
\centering
\subfigure[Sample shots of the dataset that is obtained by combining frames from two different sequences. The object enclosed in the ellipse is the object to be tracked.]{\includegraphics[height=0.15\linewidth, width=0.80\linewidth]{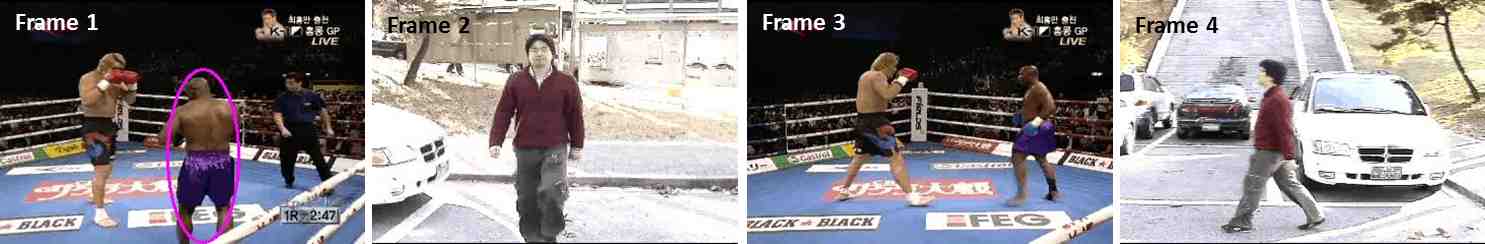}
\label{fig:LO3}}
\subfigure[Sample detections by the A-WLMC tracking. A-WLMC tend to tracked the object inaccurately once it has lost or missed tracked of the object as shown from Frame 79 onwards.]{\includegraphics[height=0.2\linewidth, width=0.85\linewidth]{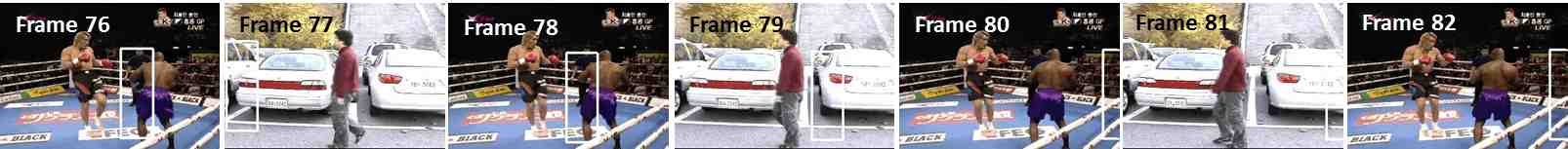}
\label{fig:LO4}}
\subfigure[Sample detections by the SwaTrack tracking. In Frame 77, since the object-of-interest does not appear in the frame, inaccurate tracking happens. However, the SwaTrack is able to recover its tracking at the following frame, Frame 78.]{\includegraphics[height=0.2\linewidth, width=0.85\linewidth]{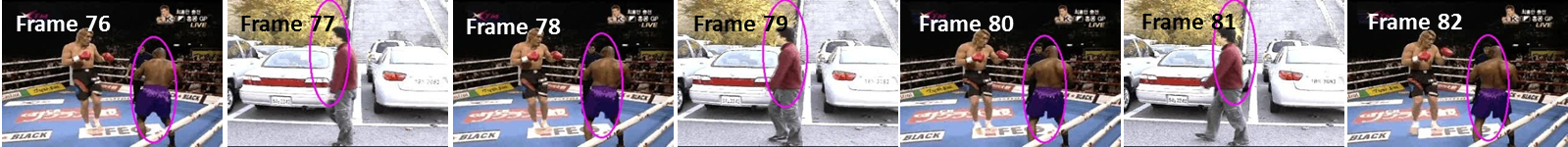}
\label{fig:LO5}}
\caption{Sample outputs to demonstrate the capability to recover from incorrect tracking.}
\label{fig:InaccurateTracking}
\end{figure}

Secondly, we simulate another challenging scenario in which incorrect tracking is most likely to happen by sampling frames from 2 different datasets as shown in Fig.\ref{fig:InaccurateTracking} (a). The frames in the Boxing sequence are combined in an alternative manner with the frames from the Youngki sequence. In this combined sequence, the object-of-interest which is highlighted in the ellipse in Frame 1 of in Fig. \ref{fig:InaccurateTracking} (a) tend to disappear from one frame and re-appear in the subsequent frame interchangeably. From the qualitative results shown in Fig. \ref{fig:InaccurateTracking} (b), we observe that the A-WLMC tracking \cite{Kwon2012} is not robust and does not cope well with inaccurate tracking. When the object-of-interest disappear from the scene (i.e. Frame 77), the A-WLMC gives an erroneous estimation of the object. In the subsequent frame, where the object re-appears, the A-WLMC has difficulty to recover from its tracking such as shown in Frame 78 where the estimation does not fit the actual position of the object accurately. In the subsequent frames, the A-WLMC tend to continuously missed tracked of the object. Although the sampling efficiency in the A-WLMC adopts a more efficient proposal distribution as compared to the standard PF, it is still subjected to a certain degree of trapped in local optima. Furthermore, the A-WLMC utilizes the information of historical samples for intensive adaptation, thus requiring more frames information to recover from inaccurate tracking. The proposed SwaTrack on the other hand, is observed to work well in this experiment, where minimal frame is required to recover from erroneous tracking. As shown in Fig. \ref{fig:InaccurateTracking} (c), the SwaTrack is able to track the object accurately when the object appear or re-appears in the scene (as shown in the even frame number). The inaccurate tracking in the odd frame number is reasonable as the object does not appear in those scenes. This is made possible due to the information exchange and cooperation between particles in a swarm that provide a way to escape the local optima and reach the global maximum; leading to and optimised proposal distribution. 

\begin{figure}[htbp]
\centering
\subfigure[Sample detections from A-WLMC on reduced image size. A-WLMC has a high tendency to lost track of the object when it moves abruptly, and demonstrate continuous inaccurate tracking such as shown in Frame 279-284. Note that for similar frames, MCMC is able to track the object accurately when the image size is larger. Number of iterations = 600, particles = 600.]{\includegraphics[height=0.15\linewidth, width=0.8\linewidth]{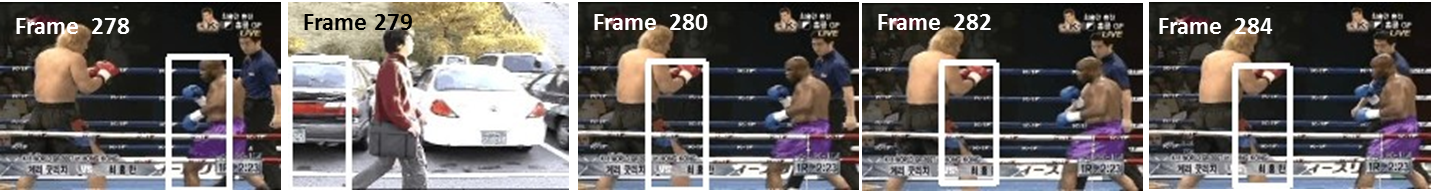}
\label{fig:LO6}}
\subfigure[Sample detections from SwaTrack on reduced image size. SwaTrack produces consistent tracking as compared to PF and A-WLMC, regardless of the size of object. Number of iterations = 30, particles =20.]{\includegraphics[height=0.15\linewidth, width=0.8\linewidth]{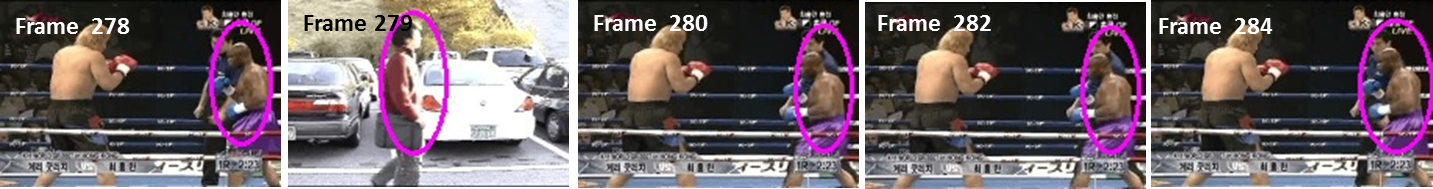}
\label{fig:LO7}}
\caption{Qualitative Results: Comparison between A-WLMC and Our Proposed Method in term of different image size.}
\label{fig:ResizeTracking}
\end{figure}

{\bf Sensitivity to Object Size} We further tested the proposed SwaTrack, PF \cite{Yan,Maggio} and A-WLMC \cite{Kwon2012} on resized sequences of similar data to simulate scenario in which the object size is smaller. Thus, the initial frame size of 360x240 is reduced into half, to 180x120 pixels. From our observations, the SwaTrack is the least sensitive towards the size of object-of-interest, while the detection accuracy of the A-WLMC is reduced as the size of object gets smaller. This is due to the robustness of the optimised sampling in SwaTrack as compared to the least robust method of rejection and acceptance as proposed in the A-WLMC. The overall detection accuracy of the proposed SwaTrack remain at an average of 90\% regardless of the object's size whereas the detection accuracy of PF and A-WLMC decrease significantly by more than 25\% when the object's size decreases. Sample output is as shown in Fig. \ref{fig:ResizeTracking}.

\begin{figure}[htbp]
\begin{center} 
\includegraphics[height=0.55\linewidth, width=0.8\linewidth]{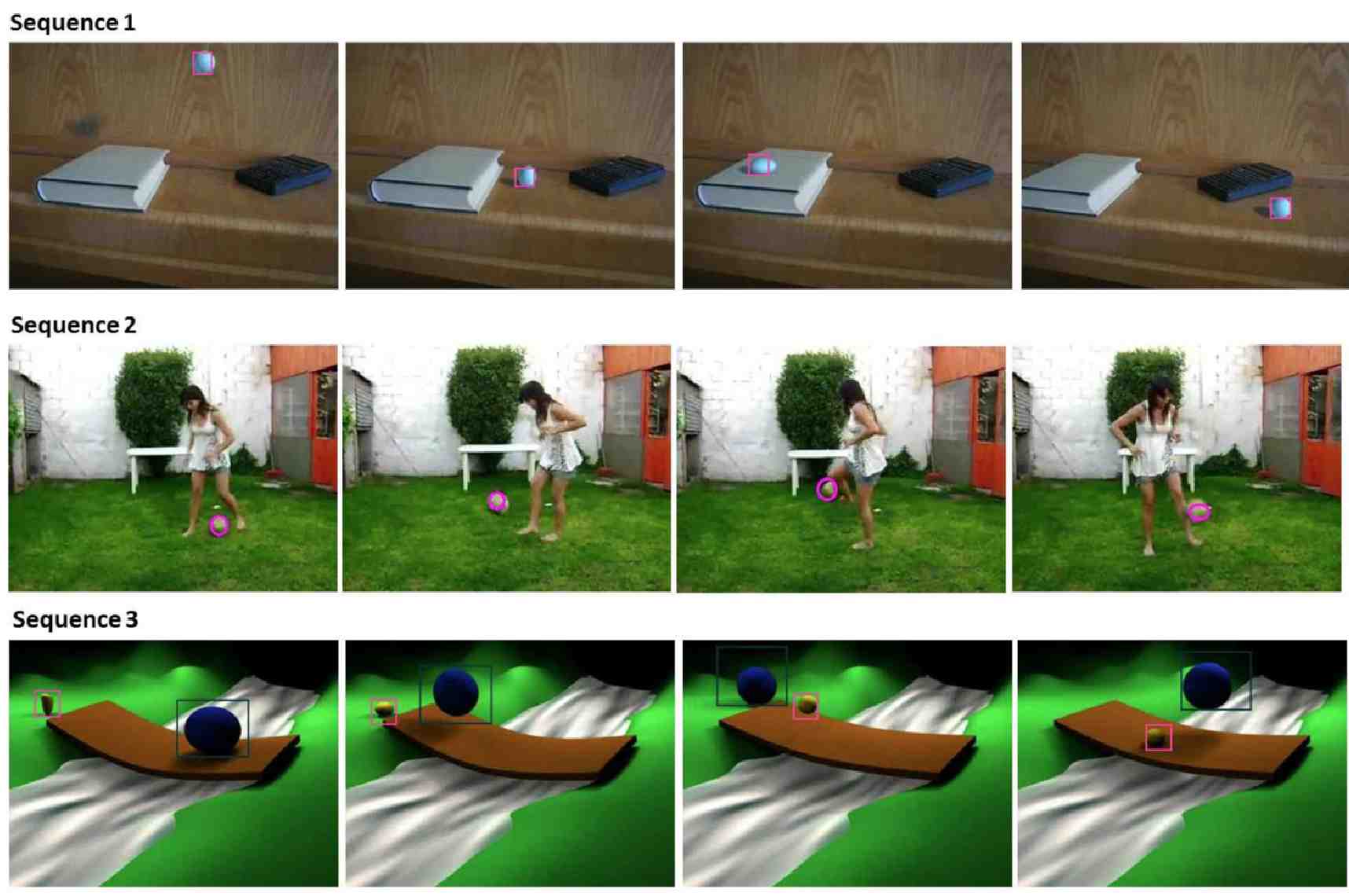}
\caption{Sample shots of tracking results on our proposed method.
} \label{fig:resultsummary02} 
\end{center}
\end{figure}

Finally, we evaluated the proposed SwaTrack on videos obtained from Youtube and the qualitative results are as depicted in Fig. \ref{fig:resultsummary02}. {\bf Abrupt motion with Inconsistent Speed:} The first aim to track a synthetic ball which moves randomly across the sequence with inconsistent speed, whilst the second sequence tracks a soccer ball which is being juggled in a free-style manner in a moving scene with a highly textured background (grass). It is observed that the SwaTrack is able to track the abrupt motion of the balls efficiently. {\bf Multiple targets:} In the third sequence, we demonstrate the capability of the proposed system to track multiple targets; two simulated balls moving at random. Most of the existing solutions are focused on single target.

\subsection{Sampling-based vs Iterative-based Solutions}

Motivated by the meta-level question prompted in \cite{Zhu} on \emph{whether there is a need to have more training data or better models for object detection}, we raise similar question in the domain of this area; will continued progress in visual tracking be driven by the increased complexity of tracking algorithms? Intuitively, an increase in the number of samples in sampling-based tracking methods such as PF and MCMC would increase the tracking accuracy. One may also argue that the additional computational cost incurred in the iterative nature of the proposed SwaTrack and MCMC would complement the higher number of particles required by the PF. Thus, in order to investigate if these intuitions hold true, we perform experiments using an increasing number of samples and iterations. We then observe the behaviors of PF and SwaTrack in terms of accuracy and processing time with the increase in complexity. PF is chosen in this testing as it bears close resemblance to the proposed SwaTrack algorithm in which a swarm of particles are deployed for tracking. 

\subsubsection{Sample of Particles vs Accuracy}

\begin{figure}[htbp]
\centering
\subfigure[PF]{{\includegraphics[height=0.3\linewidth, width=0.4\linewidth]{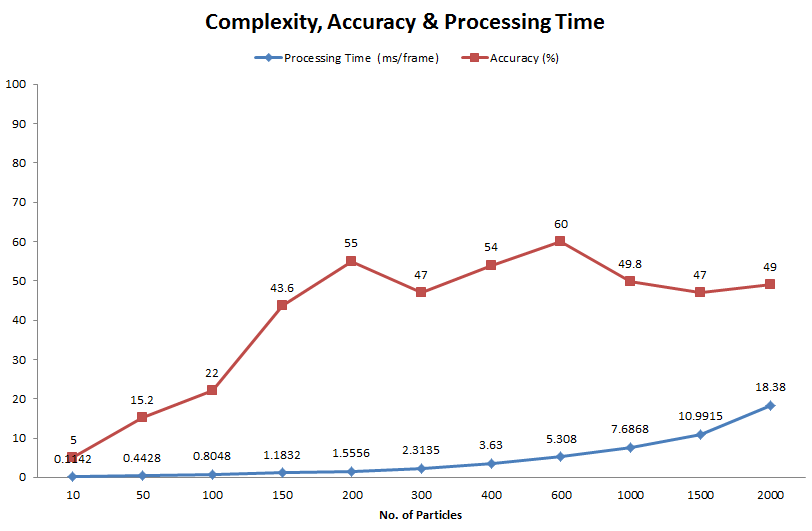}}}
\subfigure[SwaTrack]{{\includegraphics[height=0.3\linewidth, width=0.4\linewidth]{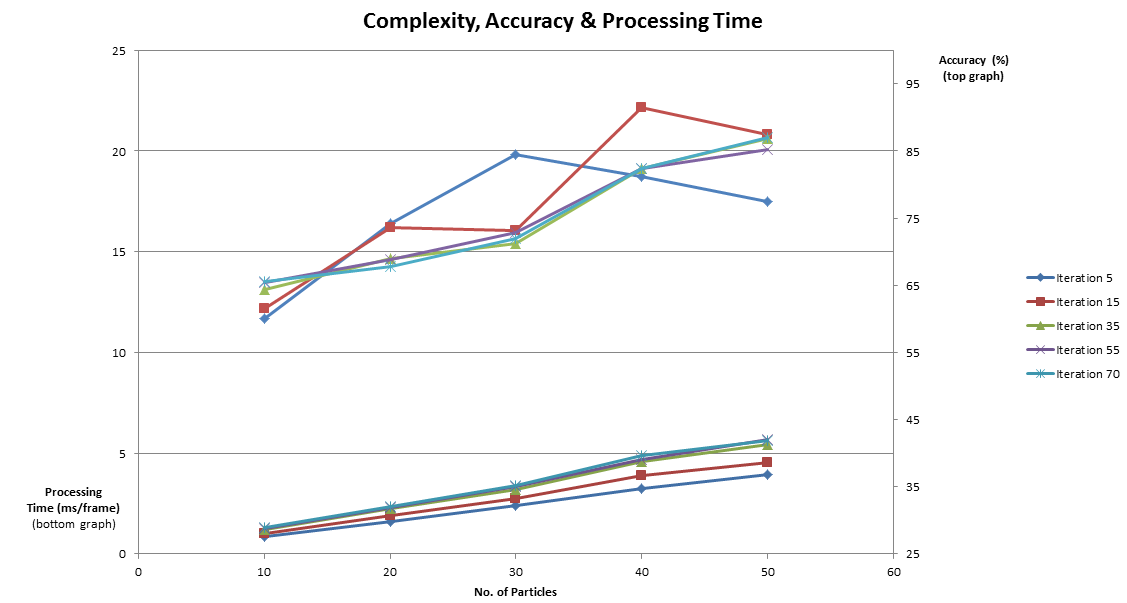}}}
\caption{A comparison between PF and SwaTrack in terms accuracy vs different number of samples and accuracy vs different number of samples and iteration}
\label{fig:PFAverage}
\end{figure}

\begin{figure}[htbp]
\centering
\subfigure[Sequence 1.]{\includegraphics[height=0.3\linewidth, width=0.4\linewidth]{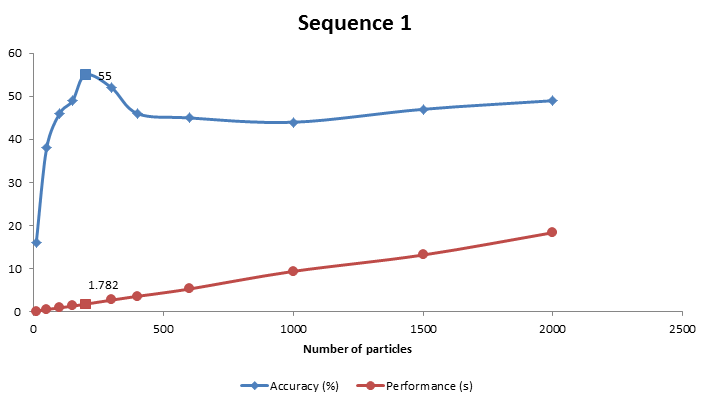}
\label{fig:LO8}}
\subfigure[Sequence 2.]{\includegraphics[height=0.3\linewidth, width=0.4\linewidth]{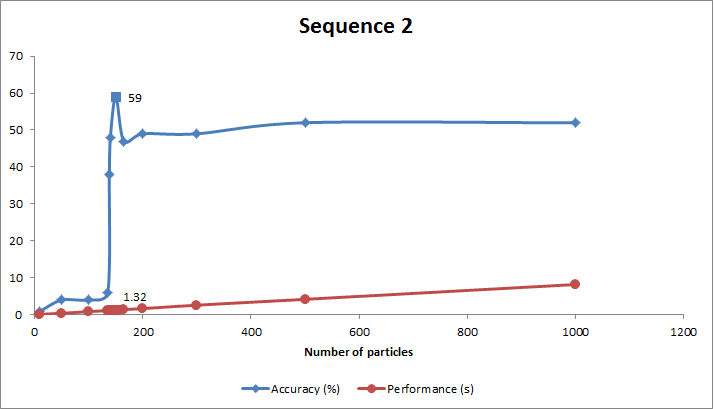}
\label{fig:LO9}}
\caption{The accuracy and performance of PF with different number of samples for $TableT1-2$.}
\label{fig:PFSequence}
\end{figure}

{\bf Particle Filter:}
In the PF algorithm, we vary the number of samples or particles (i.e. 50, 100, $\cdots$, 2000) used throughout the sequence to determine the statistical relationship between number of samples and performance. We gauge the performance by the detection accuracy (\%) and processing time (in milliseconds per frame). The average performance across all 5 datasets (TableT) are as shown in Fig. \ref{fig:PFAverage}a. Sample of the performance for sequence $TableT1$ and $TableT2$ are as shown in Fig. \ref{fig:PFSequence}.

The results demonstrate that the number of particles used in PF is correlated to the detection accuracy; where the increase in the number of particles tends to increase the accuracy. Similarly, the average time taken also increases exponentially as the number of particles used in PF grows. This alludes the fact that as the number of particles increase, the estimation processes which include object representation, prediction and update also multiply. However, it is observed that PF reaches plateau after hitting the optimal accuracy, after which any increase in the number of particles will either have a decrease in accuracy or no significant improvement. From Fig. \ref{fig:PFAverage}, we can see the detection accuracy decreases after the optimal solution, which is given when the number of particles are 600. Our findings provoke the underlying assumption that the increase of number of particles will lead to an increase in the accuracy. Thus, we raise the question of whether complex (in this context the complexity is proportional to the number of particles deployed) tracking methods are really necessary? Also, the best parameter configurations may differ from one sequence to another due to the different motion behaviour portrayed by the object in each sequence. For example, in Fig. \ref{fig:PFSequence}(a), the optimal setting is 250 particles which produces detection accuracy of 55\% and takes 1.78 seconds of processing time. Whilst the second sequence has a different optimal setting of 150 particles as shown in Fig. \ref{fig:PFSequence}(b). This advocates the notion as in \cite{Garcia} that \emph{motion models indeed only work for sometimes}. 

\begin{figure}[htbp]
\centering
\subfigure[Sequence 1.]{\includegraphics[height=0.3\linewidth, width=0.4\linewidth]{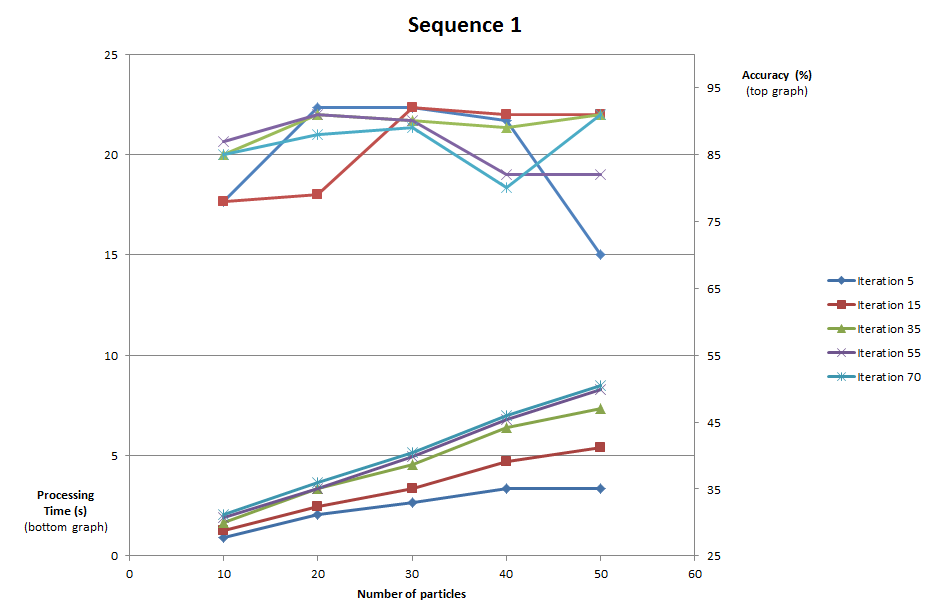}
\label{fig:LO10}}
\subfigure[Sequence 2.]{\includegraphics[height=0.3\linewidth, width=0.4\linewidth]{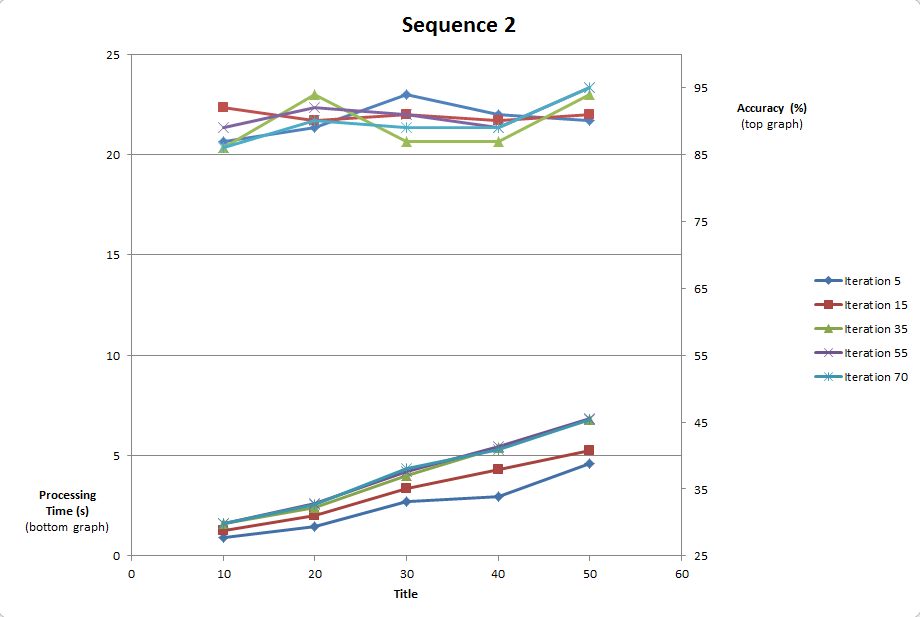}
\label{fig:LO11}}
\caption{The accuracy and performance of SwaTrack with different number of samples and iterations for $TableT1-2$.}
\label{fig:APSOSequence}
\end{figure}

{\bf SwaTrack:} Similarly, we perform the different parameter settings test on the proposed SwaTrack algorithm and the average results are demonstrated in Fig. \ref{fig:PFAverage}b, while Fig. \ref{fig:APSOSequence} illustrate the results for $TableT1$ and $TableT2$. In addition to the number of particles used in PF tracking, the proposed SwaTrack has an additional influencing parameter, the maximum number of iterations. We vary the number of particles against the number of iterations for fair evaluation. As illustrated in the left y-axis of the chart (bottom graph), we can see that the average processing time increases as the number of iterations increase. However, the increment seems to be valid until it reaches a maximal value; in which any increase in the iterations would not incur much difference in its processing time. Notice that the processing time for different higher number of iterations (55 \& 70) tend to overlap with one another, demonstrating minimal increase in processing time as the number of iterations grows. This is due to the optimisation capability of the proposed SwaTrack to terminate its search upon convergence, regardless of the defined number of iterations. This is particularly useful in ensuring efficient search for the optimal solution, with minimal number of particles. As for the detection accuracy, we can see that in general the average accuracy of the proposed SwaTrack is higher than PF, with an average accuracy of 92.1\% in the first sequence as shown in Fig. \ref{fig:APSOSequence}a. The sudden decrease in accuracy for SwaTrack tracking with 70 number of iterations as shown in Fig. \ref{fig:APSOSequence}a may be due to the erratic generation of random values in C++ implementation. This behaviour is not observed in other sequences, where their detection accuracy is consistent across frames. Thus far, we take an average result for each test case over 10 runs to ensure reliable results and we believe that with a higher number of runs, we would be able to obtain unbiased results without outliers. In summary, the results further validate our findings that the proposed SwaTrack is able to achieve better accuracy as compared to PF, whilst requiring only about 10\% of the number of samples used in PF with minimal number of iterations. This is made possible by an iterative search for the optimal proposal distribution, incorporating available observations rather than making strict assumptions on the motion of an object. Thus, we believe that the findings from our study create prospects for a new paradigm of object tracking. Again, we raise the question \emph{if there is a need to make complex existing tracking methods by fusing different models and algorithms to improve tracking efficiency? Would simple optimisation methods be sufficient? }

\subsubsection{Number of Samples vs Processing Time}

\begin{figure}[htbp]
\begin{center} 
\subfigure[PF]{{\includegraphics[height=0.3\linewidth, width=0.4\linewidth]{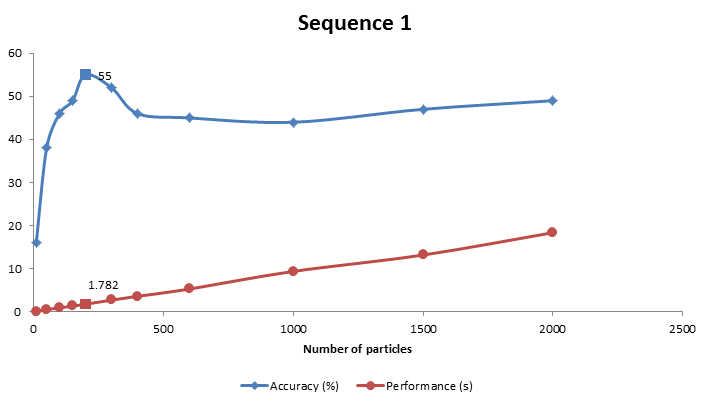}}}
\subfigure[SwaTrack]{{\includegraphics[height=0.3\linewidth, width=0.4\linewidth]{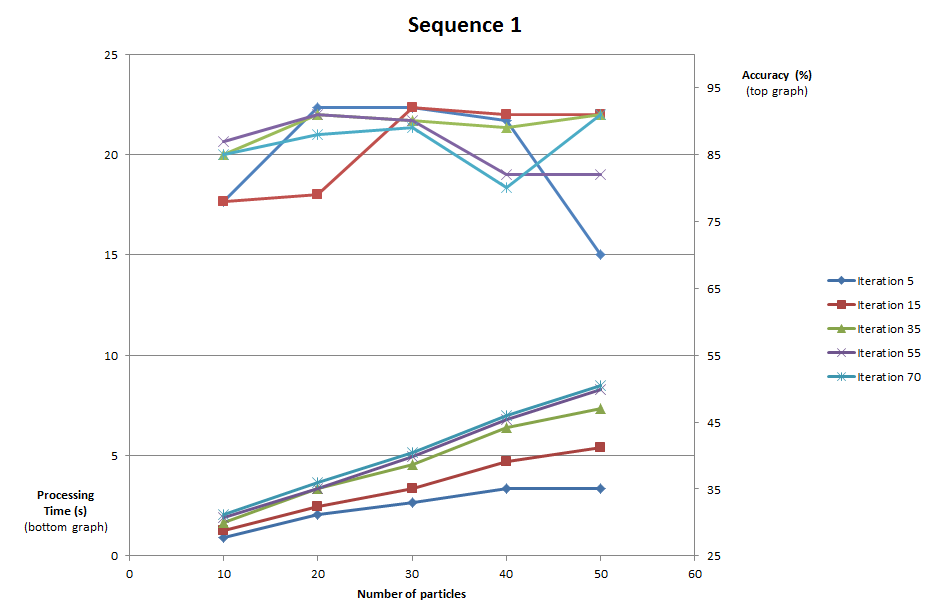}}}
\caption{ TableTennis 1: The accuracy and performance of PF/SwaTrack with different parameter settings} \label{fig:PFSequence1} 
\end{center} 
\end{figure}

\begin{figure}[htbp]
\begin{center} 
\subfigure[PF]{{\includegraphics[height=0.3\linewidth, width=0.4\linewidth]{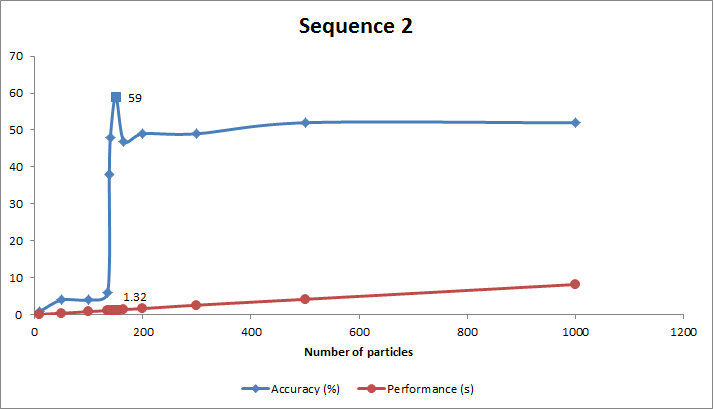}}}
\subfigure[SwaTrack]{{\includegraphics[height=0.3\linewidth, width=0.4\linewidth]{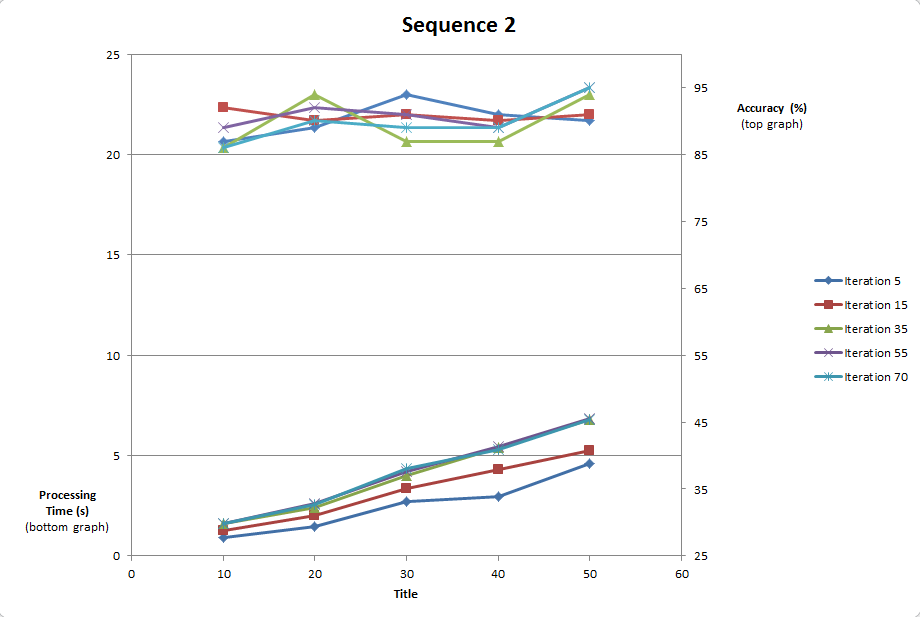}}}
\caption{ TableTennis 2: The accuracy and performance of PF with different parameter settings} \label{fig:PFSequence2} 
\end{center} 
\end{figure}

\begin{figure}[htbp]
\begin{center} 
\subfigure[PF]{{\includegraphics[height=0.3\linewidth, width=0.4\linewidth]{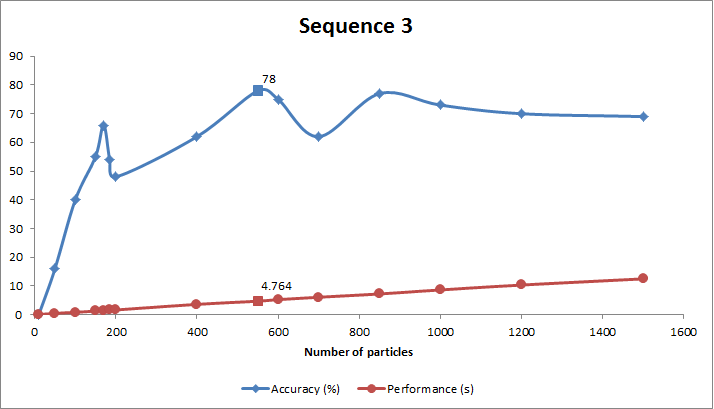}}}
\subfigure[SwaTrack]{{\includegraphics[height=0.3\linewidth, width=0.4\linewidth]{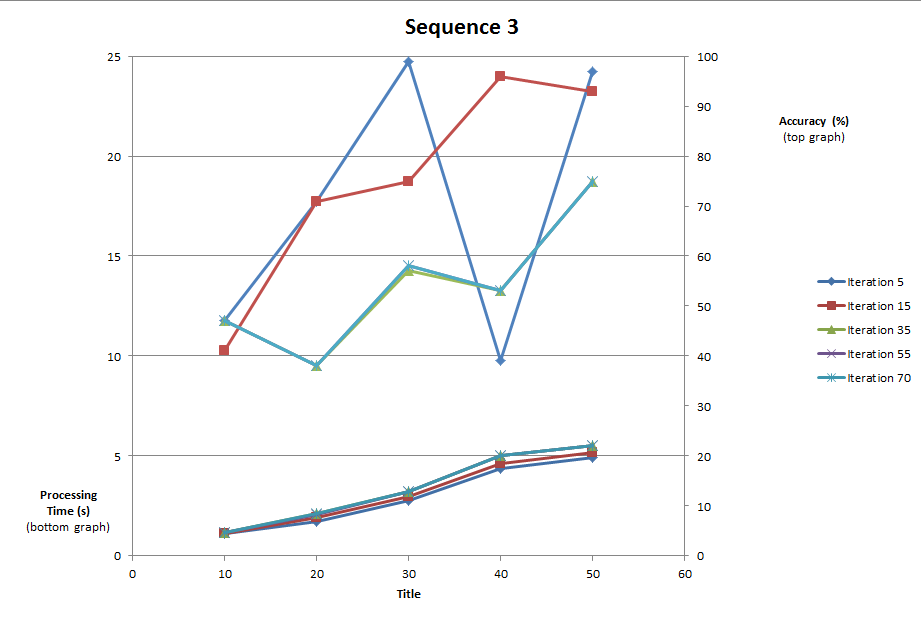}}}
\caption{ TableTennis 3: The accuracy and performance of PF with different parameter settings} \label{fig:PFSequence3} 
\end{center} 
\end{figure}

\begin{figure}[htbp]
\begin{center} 
\subfigure[PF]{{\includegraphics[height=0.3\linewidth, width=0.4\linewidth]{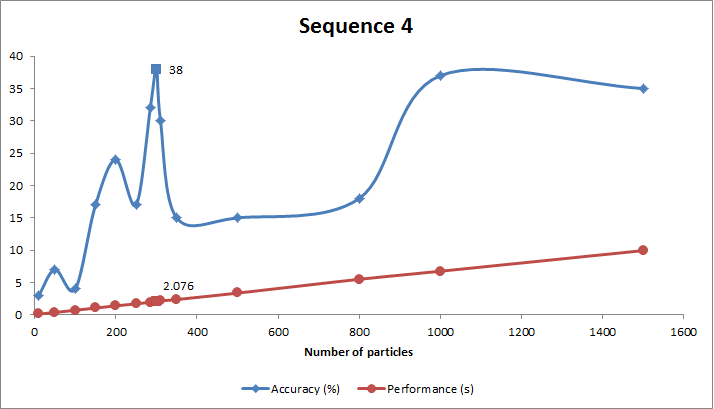}}}
\subfigure[SwaTrack]{{\includegraphics[height=0.3\linewidth, width=0.4\linewidth]{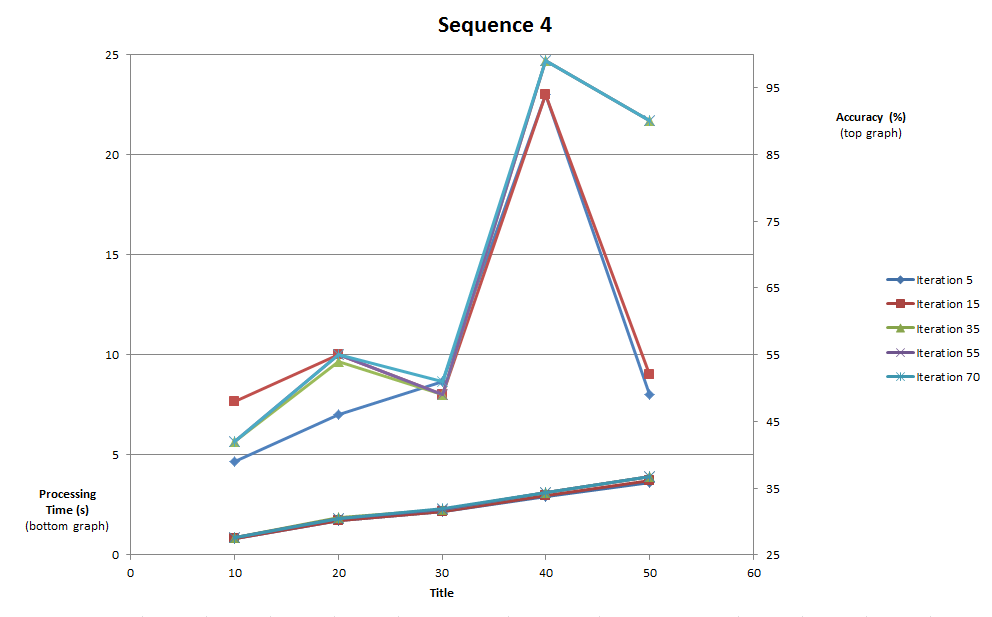}}}
\caption{ TableTennis 4: The accuracy and performance of PF with different parameter settings} \label{fig:PFSequence4} 
\end{center} 
\end{figure}

\begin{figure}[htbp]
\begin{center} 
\subfigure[PF]{{\includegraphics[height=0.3\linewidth, width=0.4\linewidth]{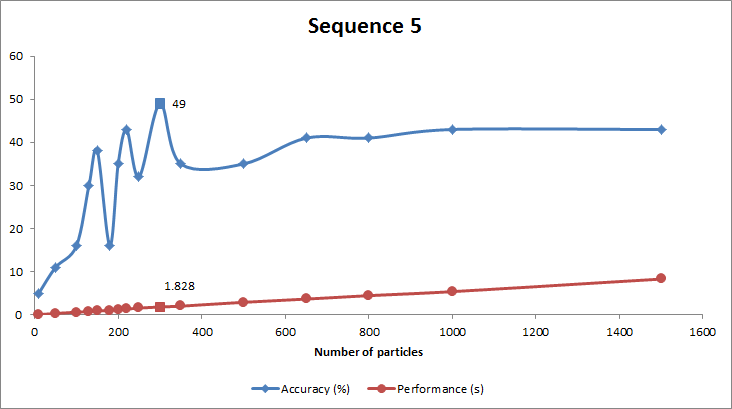}}}
\subfigure[SwaTrack]{{\includegraphics[height=0.3\linewidth, width=0.4\linewidth]{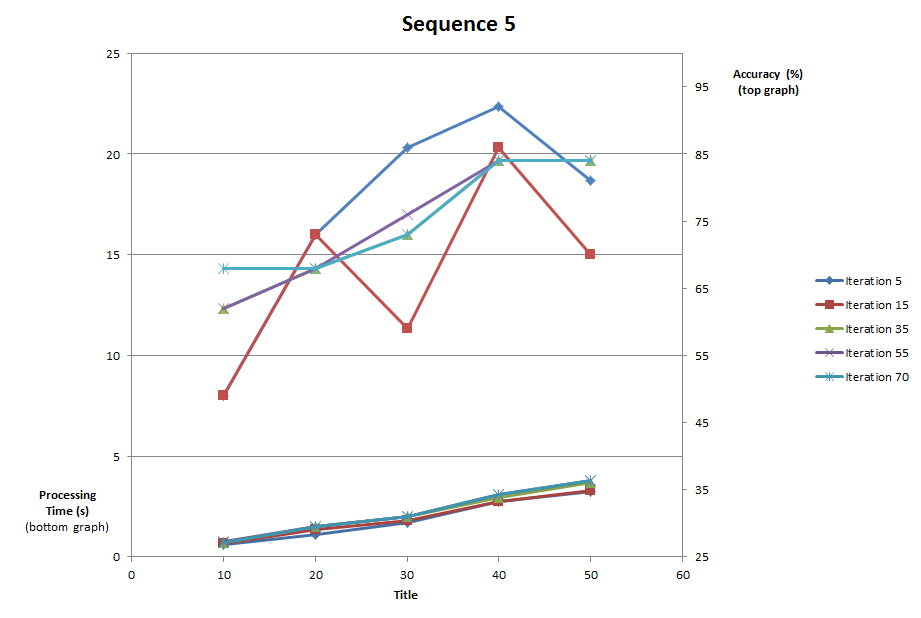}}}
\caption{ TableTennis 5: The accuracy and performance of PF with different parameter settings} \label{fig:PFSequence5} 
\end{center} 
\end{figure}

{\bf Particle Filter:}
In the PF algorith,m we vary the number of samples or particles (i.e. 50, 100, $\cdots$, 2000) used throughout the sequence to determine the statistical relationship between number of samples and detection accuracy. The lowest value of the parameter value is determined based on the minimal configuration to allow tracking while the highest value is set to the maximal configuration before it reaches a plateau detection accuracy. The detection accuracy and performance of the PF algorithm with different parameter settings are as shown in Fig. \ref{fig:PFSequence1}-\ref{fig:PFSequence5}a. The results demonstrate that the number of particles used in PF is correlated to the detection accuracy; where the increases in the number of particles tend to increase the accuracy. Similarly, the average time taken also increases as the number of particles used in PF grows. This alludes the fact that as the number of particles increase, the estimation processes which include object representation, prediction and update also multiply. However, it is observed that PF reaches a plateau detection accuracy after hitting the optimal accuracy, after which any increase in the number of particles will either have a decrease in accuracy or no significant improvement. Thus, the underlying assumption that the increase of number of particles will lead to an increase in the accuracy does not hold true. This may be due to the resampling step in most PF algorithms that is highly prone to error propagation. Also, the best parameter configurations may differ from one sequence to another due to the different motion behaviour portrayed by the object in each sequence. For example, in Fig. \ref{fig:PFSequence1}, the optimal setting is 250 number of particles which produces detection accuracy of 55\% and takes 1.78 seconds of processing time. Note that in this set of experiments, other parameters such as the mean and variance for the Gaussian distribution in PF is not optimal values as compared to the earlier experiment. A standard configuration of Gaussian white noise is used across frames. Thus, the results obtained may slightly differ. 

{\bf SwaTrack:} Similarly, we perform the different parameter settings test on the proposed SwaTrack algorithm and the results are demonstrated in Fig. \ref{fig:PFSequence1}-\ref{fig:PFSequence5}b. In addition to the number of particles used in PF tracking, the proposed SwaTrack has an additional influencing parameter, the maximum number of iterations. Thus, here we vary the number of particles against the number of iterations and obtain their detection accuracy. As illustrated in the left y-axis of the chart (bottom graph), we can see that the average processing time increases as the number of iterations increase. However, the processing time taken reaches a maximal value, where the different number of iterations require almost comparable amount of time. This can be seen by the overlapping results as shown in Fig. \ref{fig:PFSequence1}-\ref{fig:PFSequence5}b, in particular. This demonstrate that the effectiveness of the termination criteria in the proposed SwaTrack. When a global solution has been found by the entire swarm (swarm reaches convergence), the search activity terminates despite the initial setting value of the number of iterations. Also, our proposed method which automatically changes the number of iterations according to the swarm’s search quality allows a self-tuned setting of the maximum iteration number. As for the detection accuracy, we can see that in general the average accuracy of the proposed SwaTrack is higher than PF, with an average accuracy of 92.1\% in the first sequence as shown in Fig. \ref{fig:PFSequence1}b. In summary, the results further validate our findings that the proposed SwaTrack is able to achieve better accuracy as compared to PF, whilst requiring only about 10\% of the number of samples used in PF with minimal number of iterations.

\begin{figure}[htbp]
\begin{center} 
\includegraphics[height=0.4\linewidth, width=0.8\linewidth]{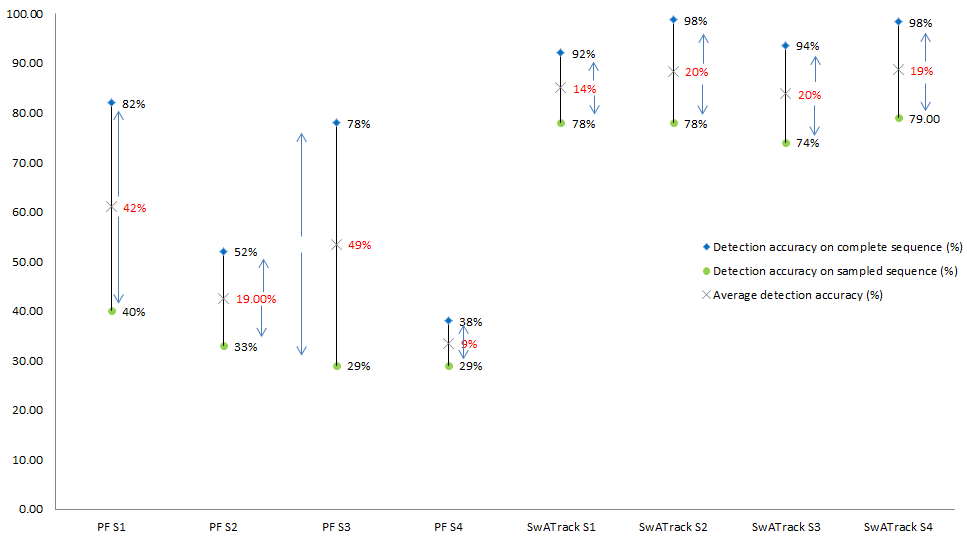}
\caption{ The detection accuracy of SwaTrack against PF during sampling for sequence 1 to 4.} \label{fig:Sampling1} 
\end{center} 
\end{figure}

\subsubsection{Sampling Strategy}
To further evaluate the robustness of the proposed algorithm as well as to understand the behaviours of other algorithms when tracking abrupt motion, we perform the sampling strategy test. In this test, we simulate the scenario of receiving inputs from the sensors with a lower frame rate by downsampling the number of frames from the test sequence; assuming the actual data are obtained at normal rate of 25 frames per second to a lower rate of 5 frames per second. 

Fig. \ref{fig:Sampling1} demonstrates the detection accuracy between the proposed SwaTrack and PF for all four sequences by down-sampling each sequence to simulate the 5 frames per second scenario. Note that the detection accuracy is determined by comparing the ground truth for the sampled frames only. However, it is observed that in general the proposed SwaTrack has better detection accuracy as compared to PF in both situations; with and without sampling. The average detection accuracy of SwaTrack for the complete sequences is about 95.5\% whereas the average for PF is about 62.5\%. During sampling, the average detection of accuracy of SwaTrack is about 77.25\% whereas PF is about 32.75\%. We can see that the detection accuracy of PF drops drastically when the frame rate decreases. This is because, in low frame rate videos, the target tends to have abrupt motion and thus, methods that assume Gaussian distribution in its dynamic motion model such as PF fail in such cases. The changes between detection accuracy on complete and sampled sequence is as indicated in red in Fig. \ref{fig:Sampling1}. SwaTrack on the other hand, copes better with low frame rate with an average accuracy of more than 70\% although there is a decrease in its efficiency. This is because, the proposed SwaTrack algorithm allows iterative adjustment of the exploration and exploitation of the swarm in search for the optimal motion model without making assumptions on the target's motion. We can thus conclude that the proposed SwaTrack algorithm is able to cope with scenarios where the frame rate is low.

\section{Conclusions}
\label{sec:conclusion}

In this paper, we presented a novel swarm intelligence-based tracker for visual tracking that copes with abrupt motion efficiently. The proposed SwaTrack optimised the search for the optimal distribution without making assumptions or need to learn the motion model before-hand. In addition, we introduced an adaptive mechanism that detects and responds to changes in the search environment to allow on the tuning of the parameters for a more accurate and effective tracking. Experimental results show that the proposed algorithm improves the accuracy of tracking while significantly reduces the computational overheads, since it requires less than 20\% of the samples used by PF. In future, we would like to further investigate the robustness of the proposed method as well as its behaviour change with the different parameter settings and sampling strategy.

\bibliographystyle{elsarticle-num}
\bibliography{myref}

\begin{thebibliography}{10}
\expandafter\ifx\csname url\endcsname\relax
  \def\url#1{\texttt{#1}}\fi
\expandafter\ifx\csname urlprefix\endcsname\relax\def\urlprefix{URL }\fi
\expandafter\ifx\csname href\endcsname\relax
  \def\href#1#2{#2} \def\path#1{#1}\fi

\bibitem{Yang}
H.~Yang, L.~Shao, F.~Zheng, L.~Wang, Z.~Song, Recent advances and trends in
  visual tracking: A review, Neurocomp. 74 (2011) 3823--3831.

\bibitem{Yilmaz}
A.~Yilmaz, O.~Javed, M.~Shah, Object tracking: A survey, ACM Comp. Surv. 38.

\bibitem{welch1995introduction}
G.~Welch, G.~Bishop, An introduction to the kalman filter (1995).

\bibitem{wan2000unscented}
E.~A. Wan, R.~Van Der~Merwe, The unscented kalman filter for nonlinear
  estimation, in: AS-SPCC, 2000, pp. 153--158.

\bibitem{Oussalah200085}
M.~Oussalah, J.~D. Schutter, Possibilistic kalman filtering for radar 2d
  tracking, Information Sciences 130~(1–4) (2000) 85 -- 107.

\bibitem{isard1998condensation}
M.~Isard, A.~Blake, Condensation—conditional density propagation for visual
  tracking, IJCV 29~(1) (1998) 5--28.

\bibitem{arulampalam2002tutorial}
M.~S. Arulampalam, S.~Maskell, N.~Gordon, T.~Clapp, A tutorial on particle
  filters for online nonlinear/non-gaussian bayesian tracking, IEEE TSP 50~(2)
  (2002) 174--188.

\bibitem{Liu2012141}
Efficient visual tracking using particle filter with incremental likelihood
  calculation, Information Sciences 195 (2012) 141 -- 153.

\bibitem{ellis2011linear}
L.~Ellis, N.~Dowson, J.~Matas, R.~Bowden, Linear regression and adaptive
  appearance models for fast simultaneous modelling and tracking, IJCV 95~(2)
  (2011) 154--179.

\bibitem{Garcia}
F.~J. Cristina Garcïa~Cifuentes, Marc~Sturzel, G.~J. Brostow, Motion models
  that only work sometimes, in: BMVC, 2012, pp. 1--12.

\bibitem{Li2009}
Y.~Li, H.~Ai, T.~Yamashita, S.~Lao, M.~Kawade, Tracking in low frame rate
  video: A cascade particle filter with discriminative observers of different
  life spans, IEEE TPAMI 30~(10) (2008) 1728--1740.

\bibitem{Kwon2008}
J.~Kwon, K.~M. Lee, Tracking of abrupt motion using wang-landau monte carlo
  estimation, in: ECCV, pp. 387--400.

\bibitem{Kwon2010}
J.~Kwon, K.~M. Lee, Visual tracking decomposition, in: CVPR, 2010, pp.
  1269--1276.

\bibitem{Kwon2012}
J.~Kwon, K.~M. Lee, Wang-landau monte carlo-based tracking methods for abrupt
  motions, IEEE TPAMI 35~(4) (2013) 1011--1024.

\bibitem{Zhang2010}
X.~Zhang, W.~Hu, X.~Wang, Y.~Kong, N.~Xie, H.~Wang, H.~Ling, S.~Maybank, A
  swarm intelligence based searching strategy for articulated 3d human body
  tracking, in: CVPRW, 2010, pp. 45--50.

\bibitem{Zhou}
X.~Zhou, Y.~Lu, J.~Lu, J.~Zhou, Abrupt motion tracking via intensively adaptive
  markov-chain monte carlo sampling, IEEE TIP 21 (2012) 789 --801.

\bibitem{Eberhart}
R.~Eberhart, J.~Kennedy, A new optimizer using particle swarm theory, in: MHS,
  1995, pp. 39--43.

\bibitem{vandenBergh2006937}
A study of particle swarm optimization particle trajectories, Information
  Sciences 176~(8) (2006) 937 -- 971.

\bibitem{zhang2008sequential}
X.~Zhang, W.~Hu, S.~Maybank, X.~Li, M.~Zhu, Sequential particle swarm
  optimization for visual tracking, in: CVPR, 2008, pp. 1--8.

\bibitem{Tong}
G.~Tong, Z.~Fang, X.~Xu, A particle swarm optimized particle filter for
  nonlinear system state estimation, in: CEC, 2006, pp. 438 --442.

\bibitem{Neri201396}
F.~Neri, E.~Mininno, G.~Iacca, Compact particle swarm optimization, Information
  Sciences 239 (2013) 96 -- 121.

\bibitem{li2009contour}
W.~Li, X.~Zhang, W.~Hu, Contour tracking with abrupt motion, in: ICIP, 2009,
  pp. 3593--3596.

\bibitem{Wong}
K.~C.~P. Wong, L.~S. Dooley, Tracking table tennis balls in real match scenes
  for umpiring applications, BJMCS 1(4) (2011) 228--241.

\bibitem{Liu}
Y.~Liu, S.~Lai, B.~Wang, M.~Zhang, W.~Wang, Feature-driven motion model-based
  particle-filter tracking method with abrupt motion handling, Opt. Eng. 51(4).

\bibitem{Zuriarrain}
I.~Zuriarrain, F.~Lerasle, N.~Arana, M.~Devy, An mcmc-based particle filter for
  multiple person tracking, in: ICPR, 2008, pp. 1--4.

\bibitem{Zhu}
X.~Zhu, C.~Vondrick, D.~Ramanan, C.~Fowlkes, Do we need more training data or
  better models for object detection?, in: BMVC, 2012, pp. 1--11.

\bibitem{zhang2010smarter}
X.~Zhang, W.~Hu, S.~Maybank, A smarter particle filter, in: ACCV, Springer,
  2010, pp. 236--246.

\bibitem{Shi}
Y.~Shi, R.~Eberhart, A modified particle swarm optimizer, in: WCCI, 1998, pp.
  69--73.

\bibitem{Kaelbling}
L.~P. Kaelbling, M.~L. Littman, A.~W. Moore, Reinforcement learning: a survey,
  JAIR 4 (1996) 237--285.

\bibitem{Yan}
F.~Yan, W.~Christmas, J.~Kittler, A tennis ball tracking algorithm for
  automatic annotation of tennis match, Sig. Proc. (2005) 619--628.

\bibitem{Maggio}
E.~Maggio, A.~Cavallaro, Accurate appearance-based bayesian tracking for
  maneuvering targets, CVIU 113~(4) (2009) 544--555.

\bibitem{Adam}
A.~Adam, E.~Rivlin, I.~Shimshoni, Robust fragments-based tracking using the
  integral histogram, in: CVPR, 2006, pp. 798--805.

\bibitem{Smith}
K.~Smith, D.~Gatica-Perez, J.~Odobez, S.~Ba, Evaluating multi-object tracking,
  in: CVPRW, 2005, pp. 36--36.

\bibitem{Everingham}
M.~Everingham, L.~Gool, C.~K. Williams, J.~Winn, A.~Zisserman, The pascal
  visual object classes (voc) challenge, IJCV 88 (2010) 303--338.

\bibitem{torralba2011}
A.~Torralba, A.~A. Efros, Unbiased look at dataset bias, in: CVPR, 2011, pp.
  1521--1528.

\end{thebibliography}
\end{document}